\documentclass[letterpaper,11pt]{memoir}
\usepackage{dgjournal}          
\usepackage{mathptmx}
\usepackage{graphics}
\usepackage[authoryear,semicolon,sectionbib]{natbib}

\counterwithout{section}{chapter}  
\maxsecnumdepth{subsubsection}

\usepackage{amsmath,amssymb,graphicx,booktabs,longtable,pdflscape,caption,titlesec,comment,dcolumn,url}
\usepackage[colorlinks=true,linkcolor=black,anchorcolor=black,citecolor=black,filecolor=black,menucolor=black,runcolor=black,urlcolor=black]{hyperref}
\newcolumntype{d}[1]{D{.}{.}{#1}}
\titleformat*{\paragraph}{\itshape}

\usepackage{xcolor}
\usepackage{centernot}
\newcommand{\expit}{\text{expit}}

\newcommand{\Exp}{\mathbb{E}}
\newcommand{\CI}{\mathrel{\perp\mspace{-10mu}\perp}}
\newcommand{\nCI}{\centernot{\CI}}

\renewcommand{\today}{\ifcase \month \or January\or February\or March\or %
April\or May \or June\or July\or August\or September\or October\or November\or %
December\fi, \number \year}

\title{{\large\scshape{Propensity score estimation using classification and regression trees in the presence of missing covariate data}}}
\author{Bas B.L. Penning de Vries\thanks{Clinical Epidemiology, Leiden University Medical Center, 2300 RC Leiden, the Netherlands}\thanks{Corresponding author: Clinical Epidemiology, room C7-107, Leiden University Medical Center, PO Box 9600, 2300 RC Leiden, the Netherlands. \url{B.B.L.Penning_de_Vries@lumc.nl}}, Maarten van Smeden\thanksmark{1}, Rolf H.H. Groenwold\thanksmark{1}}
\date{\today}

\usepackage{soul}
\newcommand{\delete}[1]{{\color{black!50}{\st{#1}}}}
\newcommand{\replace}[2]{{\color{black!50}{\st{#1}#2}}}
\newcommand{\add}[1]{{\color{black!50}{#1}}}
\renewcommand{\delete}[1]{}
\renewcommand{\replace}[2]{#2}
\renewcommand{\add}[1]{#1}
\newcommand{\deleteII}[1]{{\color{red}{\st{#1}}}}
\newcommand{\replaceII}[2]{{\color{red}{\st{#1}#2}}}
\newcommand{\addII}[1]{{\color{red}{#1}}}
\renewcommand{\deleteII}[1]{}
\renewcommand{\replaceII}[2]{#2}
\renewcommand{\addII}[1]{#1}

\counterwithout{table}{chapter}

\begin{document}

\originalmaketitle

\begin{abstract}
\noindent
Data mining and machine learning techniques such as classification and regression trees (CART) represent a promising alternative to conventional logistic regression for propensity score estimation. Whereas incomplete data preclude the fitting of a logistic regression on all subjects, CART is appealing in part because some implementations allow for incomplete records to be incorporated in the tree fitting and provide propensity score estimates for all subjects. Based on theoretical considerations, we argue that the automatic handling of missing data by CART may however not be appropriate.
Using a series of simulation experiments, we examined the performance of different approaches to handling missing covariate data; (i) applying the CART algorithm directly to the (partially) incomplete data, (ii) complete case analysis, and (iii) multiple imputation.  Performance was assessed in terms of bias in estimating exposure-outcome effects \add{among the exposed}, standard error, mean squared error and coverage. Applying the CART algorithm directly to incomplete data resulted in bias, even in scenarios where data were missing completely at random. Overall, multiple imputation followed by CART resulted in the best performance. Our study showed that automatic handling of missing data in CART can cause serious bias and does not outperform multiple imputation as a means to account for missing data.
\end{abstract}

\section{Introduction}

Propensity score analysis has gained increasing popularity as means to adjust for measured confounding \citep{Rosenbaum1983,Sturmer2006}. Inference typically proceeds by stratification on the propensity score, propensity score adjustment in a regression model, inverse probability weighting (IPW) or matching based on propensity scores given measured covariates \citep{Rosenbaum1983,Austin2011intro}. It is standard practice to obtain estimates of the propensity score by a parametric (logistic) regression of the exposure on measured covariates. However, parametric models rely on assumptions about the distribution of variables in relation to one another, including the functional form and the presence or absence of interactions. If any of these are violated, covariate balance may not be attained, potentially leading to bias in making causal inferences about the exposure-outcome relation of interest \citep{Drake1993}.

It has been suggested that machine learning and data mining methods, such as classification and regression tree analysis (CART), be used to estimate the relationship between the exposure and measured covariates. 
These methods avoid making the assumptions regarding functional form and interaction as in a standard logistic regression. 
The utility of data mining methods to estimate propensity scores in complete data settings has been studied previously \citep{Setoguchi2008,Lee2010,Westreich2010,Wyss2014}. However, in practice, researchers are often faced with missing values on the measured variables. Whereas incomplete data preclude logistic regression on all subjects, some CART algorithms allow for incomplete records to be incorporated in the tree fitting and provide propensity score estimates for all subjects. The ability of CART to accommodate missing values has been described as advantageous \citep{Lee2010,McCaffrey2004,Moisen2008,Rai2017}. However, the precise impact of missing data on the performance of CART-based propensity score estimators has received little attention. The objective of this study was therefore to examine the performance of various CART-based propensity score estimation procedures in the presence of missing data. \add{Throughout, particular emphasis is placed on the causal odds ratio for the marginal effect among the exposed (or Average Effect among the `Treated', ATT) as the effect measure of interest.}

The remainder of this article is structured as follows. In Section~\ref{Theory}, we briefly review pertinent theory. Based on analytical work, we identify caveats in the handling of missing data by CART. Section~\ref{MonteCarloSimulations} describes a series of Monte Carlo simulations that were used to evaluate the performance of various approaches to handling missing data, including (i) subjecting incomplete data directly to the CART algorithm, (ii) complete case analysis, and (iii) multiple imputation. In Section~\ref{CaseStudy}, we apply and compare the approaches in a case study on the effect of influenza vaccination and mortality. We conclude with a summary and discussion of our findings in the context of the existing literature.

\section{Theory}\label{Theory}

\subsection{Propensity score analysis of complete data}

\subsubsection*{Counterfactual outcomes and estimating causal effects}

We adopt a perspective of potential or counterfactual outcomes, formal accounts of which are given for example by \cite{Neyman1935}, \cite{Rubin1974}, \cite{Holland1986}, \cite{Holland1988} and \cite{Pearl2009}.

Consider a\replace{n ordered set}{ sequence} $S=(X_1,X_2,...,X_n)$ of variables and let $\mathcal{F}=(f_{X_1},f_{X_2},...,f_{X_n})$ be a collection of functions $f_{X_j}$ that deterministically map a realisation of the predecessors $(X_i:i<j)$ of $X_j$ and of exogenous variable $\varepsilon_{X_j}$ into a realisation of $X_j$. We may write the random variable $X_j$ as follows:
\begin{align}\label{SCM}
X_j=f_{X_j}\big(f_{X_1}(X_1,\varepsilon_{X_1}),f_{X_2}(X_1,X_2,\varepsilon_{X_2}),...,f_{X_{j-1}}(X_1,X_2,...,X_{j-2},\varepsilon_{X_{j-1}}),\varepsilon_{X_j}\big).
\end{align}
For any intervention \delete{on a set of variables $T\subseteq S\{X_j\}$ stipulating that $X_i=x_i$ for $X_i\in T$}\add{setting $X_t=x_t$ for $t$ in a subset $T$ of $\{1,...,n\}\backslash\{j\}$}, the counterfactual version of $X_j$ is obtained by evaluating \add{the right-hand side of} $\eqref{SCM}$ with \replace{$X_i$ replaced by $x_i$ for all $X_i\in T$}{$X_t$ replaced by $x_t$ for all $t\in T$}.

Specifically, let $S=(W,A,Y,R)$ so that $W=f_W(\varepsilon_W)$, $A=f_A(W,\varepsilon_A)$, $Y=f_Y(W,A,\varepsilon_Y)$, and $R=f_R(W,A,Y,\varepsilon_R)$. $W$ may be thought of as a (random vector of) baseline or pre-exposure variable(s), $A$ denotes the binary exposure of interest, $Y$ the outcome, and $R$ a missing indicator vector of $W$. A subject's counterfactual outcomes $Y_0$ and $Y_1$, obtained if exposure $A$ were set possibly contrary to fact to 0 and 1, respectively, are defined such that $Y_0=f_Y(W,0,\varepsilon_Y)$ and $Y_1=f_Y(W,1,\varepsilon_Y)$.

Causal effects are readily defined in terms of counterfactual outcomes. In this article, the focus is on the causal odds ratio (OR) for the marginal effect of exposure $A$ on binary outcome $Y$ among the exposed ($A=1$)\delete{ (Average Effect among the `Treated', ATT)}, that is,
\begin{align*}
\text{OR} = \frac{\Exp[Y_1|A=1]/(1-\Exp[Y_1|A=1])}{\Exp[Y_0|A=1]/(1-\Exp[Y_0|A=1])}.
\end{align*}

Under consistency as defined by \cite{Cole2009}, $Y_1$ is equal to the observed outcome $Y$ for subjects in the exposure group. $Y_0$, on the other hand, is not observed for exposed subjects. We may, however, validly estimate the causal OR under a set of \replace{assumptions}{conditions}, which includes no interference between subjects (or Stable Unit Treatment Value Assumption, \citeauthor{Tchetgen2012}, \citeyear{Tchetgen2012}), consistency, positivity, and conditional exchangeability \citep{Lesko2017}. To simplify arguments and notation, we shall assume that all of these conditions hold, with the exception of conditional exchangeability, unless otherwise indicated. If there exists a (set of) variable(s) $Z$ such that the potential outcomes are conditionally independent of exposure status given $Z$, we may write
\begin{align*}
P(Y_0|A=1) &= \Exp[P(Y_0|A=1,Z)|A=1] \\
	&= \Exp[P(Y_0|A=0,Z)|A=1] \\
&= \Exp[P(Y|A=0,Z)|A=1],
\end{align*}
so that the causal OR may be expressed entirely in terms of observables
\begin{align*}
\text{OR} = \frac{\Exp[Y|A=1]/(1-\Exp[Y|A=1])}{\Exp\{\Exp[Y|A=0,Z]|A=1\}/(1-\Exp\{\Exp[Y|A=0,Z]|A=1\})}.
\end{align*}
$W$ satisfies the definition of $Z$ whenever $\varepsilon_Y\CI \varepsilon_A|W$. In practice, validly estimating $\Exp[Y|A=0,Z]$ may be difficult when $Z$ is multidimensional and $Y$ is rare \citep{Albert1984}. In this case, it may be desirable to summarise $Z$ in a single balancing score \citep{Rosenbaum1983}.

\subsubsection*{The propensity score}

The propensity score $e(W)$, defined as the conditional probability of exposure given covariates $W$, satisfies a number of balancing properties. First, covariate(s) $W$ and exposure $A$ are conditionally independent given the propensity score, and conditional exchangeability given covariate(s) $W$ implies conditional exchangeability given $e(W)$ \citep[Theorems 1 and 3]{Rosenbaum1983}.
Thus, the causal OR becomes \begin{align*}
\text{OR} = \frac{\Exp[Y|A=1]/(1-\Exp[Y|A=1])}{\Exp\{\Exp[Y|A=0,e(W)]|A=1\}/(1-\Exp\{\Exp[Y|A=0,e(W)]|A=1\})}.
\end{align*}
This formulation has motivated the propensity score matching approach as discussed by \cite{Rosenbaum1983}.

Balance may also be attained by inverse probability weighting (Appendix A). To simplify arguments and notation, we assume that $W$ and $Y$ take a discrete joint distribution; however, the results extend to continuous or mixed discrete/continuous distributions.
To obtain an IPW estimator of the ATT, let
$$\varphi(w,a)=\frac{\varphi^\ast(w,a)}{\mathbb{E}[\varphi^\ast(W,A)|A=a]},\qquad \varphi^\ast(w,a)=I(a=1)+I(a=0)\frac{e(w)}{1-e(w)},$$
for realisations $w$ of $W$ and $a$ of $A$, where $I$ denotes the indicator function taken the value 1 if the argument is true and 0 otherwise.
Weighting by $\varphi$ yields independence between covariate(s) $W$ and $A$; that is, for all $w$, $$\varphi(w,0)\Pr(W=w|A=0)=\varphi(w,1)\Pr(W=w|A=1).$$ Also, conditional exchangeability given $W$ implies exchangeability following weighting by $\varphi$; that is, if $(Y_0,Y_1)\CI A|W=w$ for all $w$, then \begin{align*}\sum_w\varphi(w,0)\Pr(Y_0&=y_0,Y_1=y_1,W=w|A=0)\\&=\sum_w\varphi(w,1)\Pr(Y_0=y_0,Y_1=y_1,W=w|A=1)\end{align*}
for all $y_0,y_1$. Thus, the causal OR becomes \begin{align*}
\text{OR} &= \frac{\sum_w\varphi(w,1)\Pr(Y=1,W=w|A=1)}{\big\{1-\sum_w\varphi(w,1)\Pr(Y=1,W=w|A=1)\big\}}\\&\qquad\Big/\frac{\sum_w\varphi(w,0)\Pr(Y=1,W=w|A=0)}{\big\{1-\sum_w\varphi(w,0)\Pr(Y=1,W=w|A=0)\big\}}.
\end{align*}
In words, this means that the causal odds ratio is equal to the crude odds ratio of the ATT in the (pseudo-)population that is obtained by weighting each observation by $\varphi$.

\subsubsection*{Ensemble CART methods in the absence of missing data}

We will now briefly describe how CART can be applied to estimate the propensity score. Detailed information can be found elsewhere \citep{McCaffrey2004,Breiman1996,Ridgeway1999,Breiman2001,Elith2008,Hastie2009}. CART is a type of supervised learning task that entails finding a set of rules, subject to constraints, that partition the data into regions based on the input data (covariates) such that within regions, target values (e.g., exposure levels) meet a desirable level of homogeneity. Typically, a tree is built in a recursive manner by splitting the dataset into increasingly homogeneous subsets and choosing the splitting rule at each step or node that best splits the data further, with `best' referring to the greatest improvement in terms of some homogeneity metric, such as the Gini index \citep{Therneau2017}. Ensemble techniques by definition fit more than one tree to the data and combine them to form a single predictor of `the outcome' (in the case of propensity score, the assigned exposure). The aim of ensemble techniques is to enhance performance and reduce issues of overfitting by a single tree \citep{Elith2008,Moisen2008,Hastie2009}. We focus here on two popular CART ensemble methods, namely boostrap aggregated (bagged) CART and boosted CART.

\paragraph*{Bootstrap aggregated CART}~\\\noindent
Bagged CART involves drawing bootstrap samples form the original study sample \citep{Breiman1996}. A CART tree is formed in each bootstrap sample, yielding multiple predictors of the target variable. For each subject, the final prediction is formed by the average or majority vote across all predictors. In the context of propensity scores, the prediction of a single tree for any given subject may be defined as the proportion of exposed subjects among those individuals that are assigned to the same region by the given tree. The final propensity score is the average of the predictions across all bootstrap samples. Propensity score matching may then be thought of as matching exposed subjects to unexposed subjects from the same or `nearby' region.

\paragraph*{Boosted CART}~\\\noindent
Boosted CART is related to bagged CART in the sense that it is an ensemble method; multiple trees are fit and merged to form a single predictor. With boosted CART, trees are fit in a forward, stagewise procedure. In boosting, trees are fit iteratively to the data such those observations whose observed exposure levels are poorly predicted by the predictor of the previous iteration receive greater weight at the
current iteration \citep{Ridgeway1999,Elith2008}. Some implementations construct trees using data splits aimed not at achieving homogeneity of the exposure values themselves, but at achieving homogeneity of prediction error of the estimator obtained in the previous step \citep{McCaffrey2004,Elith2008}. With each iteration, a new predictor is formed by making adjustments to the predictor obtained in the previous step. The final predictor is constructed with contributions from all trees.

\subsection{Ignorable missing data and generalised propensity scores}

In this section, we briefly review the concept of ignorable missing data, and discuss a generalisation of the propensity score which allows for missing data as well as strategies to incorporate missing data directly in the CART fitting. For certain CART algorithms (in our case boosted CART), the inherent missing data strategy yields estimates of the generalised propensity score.

\subsubsection*{Ignorable missing data}

Suppose $W=(W_1,W_2,...,W_p)$ and $R=(R_1,R_2,...,R_p)$ are random vectors of size $p$ such that for $j=1,2,...,p$, $R_j=0$ if $W_j$ is missing and $R_j=1$ if $W_j$ is observed. Following \cite{Rubin1976}, define the extended random vector $V=(V_1,V_2,...,V_p)$ with range to include the special value $*$ to indicate a missing datum: $V_j=W_j$ if $R_j=1$ and $V_j=*$ if $R_j=0$. Let ${v}$ be a particular sample realisation of $V$, so that each $v_j$ is either a known quantity or $*$. These values imply a realisation for the random variable $R$, denoted ${r}$. For notational convenience, we write $W=(W^\mathrm{obs},W^\mathrm{mis})$ and $V=(V^\mathrm{obs},V^\mathrm{mis})$ to indicate that each may be partitioned into two vectors corresponding to all $j$ such that ${r}_j=1$ for observed data and ${r}_j=0$ for missing data. It is important to note that these partitions are defined with respect to ${r}$, the observed pattern of missing data. Given a realisation $r$ of $R$, and provided that $A,Y$ are observed, covariate data are said to be missing at random (MAR) if $\Pr(R=r|W^\mathrm{obs},W^\mathrm{mis}=u,A,Y)$ and $\Pr(R=r|W^\mathrm{obs},W^\mathrm{mis}=u',A,Y)$ are the same for all $u,u'$ and at each possible value of the \replace{missing data}{parameter} vector $\phi$\replace{, which encodes the missing data mechanism}{ that fully characterises the missing data mechanism} \citep{Rubin1976}. If in addition to MAR, the parameter $\phi$ is distinct\add{, in the sense of \cite{Rubin1976},} from the \replace{complete data parameter, denoted $\theta$, in the sense of \mbox{\cite{Rubin1976}}}{vector $\theta$ that parameterises the distribution of the data that we would have based inference on had there been no missingness, then} missing data is said to be ignorable\delete{,} and it is not necessary to consider the missing data or the missing data mechanism in making inferences about $\theta$ \citep{Rubin1976,Rubin1987,Schafer1997}. Thus, if the missing data mechanism is ignorable, one may validly model the complete data to create imputations for the missing data \citep{Rubin1987,Buuren2012}.

\subsubsection*{The generalised propensity score}

The generalised propensity score $e^\ast(V)$ is defined as the conditional exposure probability given the extended covariate vector $V$ \citep{D'Agostino2000}. That is, \begin{align*}
e^\ast(V) &=\Pr(A=1|W^\mathrm{obs},R) \\ &=\sum_w\Pr(A=1|W,R)\Pr(W^\mathrm{mis}=w|W^\mathrm{obs},R).\end{align*} Using the same argumentation to establish the balancing properties of the usual propensity score, it can be shown that the generalised propensity score has the same balancing properties with respect to $V$ as the usual propensity score has with respect to $W$. Thus, the observed covariate data and missingness information and exposure $A$ are conditionally independent given the generalised propensity score, and conditional exchangeability given the extended covariate(s) $V$ implies conditional exchangeability given the generalised propensity score $e^\ast(V)$.

To obtain an IPW estimator of the ATT, let
$$\gamma(v,a)=\frac{\gamma^\ast(v,a)}{\mathbb{E}[\gamma^\ast(V,A)|A=a]},\qquad \gamma^\ast(v,a)=I(a=1)+I(a=0)\frac{e^\ast(v)}{1-e(v)},$$
for realisations $v$ of $V$ and $a$ of $A$,
Then, weighting by $\gamma$ renders $V$ independent of $A$; that is, for all $v$, $$\gamma(v,0)\Pr(V=v|A=0)=\gamma(v,1)\Pr(V=v|A=1).$$ Also, conditional exchangeability given $V$ implies conditional exchangeability following weighting by $\gamma$; that is, if $(Y_0,Y_1)\CI A|V$, then \begin{align*}\sum_{v}\gamma(v,0)\Pr(Y_0=y_0&,Y_1=y_1,V=v|A=0)\\&=\sum_{v}\gamma(v,1)\Pr(Y_0=y_0,Y_1=y_1,V=v|A=1)\end{align*}
for all $y_0,y_1$.

Importantly, the propensity score $e(W)$ need not equal the generalised propensity score $e^\ast(V)$. That is, given the observed covariate data, the unobserved covariate data need not provide the same information about exposure allocation as does the missing data pattern. In addition, neither covariate balance given the propensity score ($W\CI A|e(W)$) nor balance of the observed data and missingness information given the generalised propensity score ($V\CI A|e^\ast(V)$) generally implies covariate balance given the generalised propensity score ($W\CI A|e^\ast(V)$).

More crucially perhaps, conditional exchangeability given the generalised propensity score is not guaranteed even if conditional exchangeability given the usual propensity score holds or the generalised propensity score balances both observed and unobserved covariate data (i.e., neither $(Y_0,Y_1)\CI A|e(W)$ nor $W\CI A|e^\ast(V)$ nor both imply that $(Y_0,Y_1)\CI A|e^\ast(V)$; see Appendix B for an example).

This suggests that it is not generally desirable to distribute across exposure groups both the observed data and the missingness information by adjusting for the generalised propensity score. However, there are situations conceivable in which it is appropriate to base inference on the generalised rather than the usual propensity score. Until now, we have assumed an ordering of the variables in which the outcome $Y$ precedes $R$, the missingness pattern of $W$. Consequently, $Y$ was defined as a function $f_Y$ of $W$, $A$ and exogenous variable $\varepsilon_Y$ and not of $R$. Consider now a setting where $S=(W,R,A,Y)$ so that $R$ forms a predecessor of $A$ and $Y$ (and, therefore, a potential common cause of $A$ and $Y$). Then, if exchangeability can be attained by conditioning on $e^\ast(V)$, conditional exchangeability given $e(W)$ need not hold (see Appendix C for an example).

Thus, the choice between adjustment for the generalised versus the usual propensity score should ideally rest on the relative extent to which conditional exchangeability holds given the generalised versus the usual propensity score. In practice, it is not possible to estimate directly the true propensity score when covariate data are missing \citep{Rosenbaum1983,D'Agostino2000,Cham2016}. However, under ignorability of missing data, one may `recover' the unobserved data, e.g., via multiple imputation \citep{Rubin1987,Buuren2012}, prior to estimating propensity scores. Henceforth, we assume that exchangeability can be attained by conditioning on the complete covariate data or, therefore, the usual propensity score, if data were not missing. We also assume that missing data is ignorable.

\subsubsection*{Applying CART to incomplete data}

\paragraph*{Bootstrap aggregated CART}~\\\noindent
In this study, we used bagged CART as implemented in the R package \texttt{ipred} \citep[version 0.9-6]{ipred2017}. This implementation allows for missing data by first evaluating homogeneity at a given node among only those observations whose candidate splitting variable is observed. Once the splitting variable and split point have been decided, the algorithm uses a surrogate splits approach to classify records whose splitting variable is missing based on the other variables included in the tree fitting \citep{Therneau2017}.

The bagged CART algorithm replaces missing confounder values without regard of the outcome or exposure status. As a result, any two subjects whose covariate data are identical, except possibly for the missing covariate, would be allocated to the same covariate region by any given tree. However, subjects within a given region need not be exchangeable. In fact, systematic differences in the outcome of the causal model ($Y$) between exposed and unexposed subjects may be in part attributable to the missing covariate (confounder). As such, even under completely at random missingness (MCAR), we would expect propensity score matching or IPW based on bagged CART to yield bias in the direction of confounding by the missing covariate.

\paragraph*{Boosted CART}~\\\noindent
An implementation of boosted CART to estimate propensity scores is available in the R package \texttt{twang} \citep[version 1.5]{twang2017}. This implementation allows for incomplete records to be incorporated in the tree fitting by regarding missingness as a special covariate level and assigning to a given (non-terminal) node three child nodes; one to which any individual is allocated whose splitting variable is missing, one for observed values that exceed some threshold, and one for the remainder. That is, rather than modelling the relationship between exposure and covariates, an attempt is made to model the association between exposure on the one hand and observed covariate data and missingness information on the other hand, and, therefore, to construct scores that balance the missingness across the matched or weighted exposure groups. In other words, the algorithm represents an estimator of the generalised propensity score.

While boosted CART may be successful at distributing missingness rates across exposure groups, it makes no attempt at distributing the unobserved values. If the partially observed covariate represents a confounder, systematic differences across exposure groups may persist after propensity score matching or IPW based on the generalised propensity score. As such, under MCAR, we would expect boosted CART to yield a propensity score matching or IPW estimator that is biased in the direction of confounding by the partially observed covariate. When missingness is MAR dependent on the outcome, boosted CART tends to render exposure groups more comparable in terms of the outcome and, therefore, attenuate the apparent exposure-outcome effect.

\paragraph*{Bias when applying CART to incomplete data}~\\\noindent
In summary, above, we argued that using either boosted CART or bagged CART to estimate propensity scores may yield a biased estimator of the causal ATT, when applying the CART algorithm directly to the (partially) incomplete data. In bagged CART, missing confounder values are replaced, yet this procedure may not be appropriate, since exposure and outcome status are ignored in this process. Boosted CART, on the other hand, balances observed covariate values as well as missing indicator values. Since the latter may depend on the outcome (under the assumption of ignorability), boosted CART potentially balances outcome values too, yielding a biased estimator of the causal effect.

\section{Monte Carlo simulations}\label{MonteCarloSimulations}

We now describe a simulation study in which we evaluated the performance of CART-based propensity score matching and IPW in the presence of missing confounder data.

\label{Methods}
\subsection{Methods}
\subsubsection*{Simulation structure}
\begin{figure}[b]\centering
\includegraphics[scale=.6]{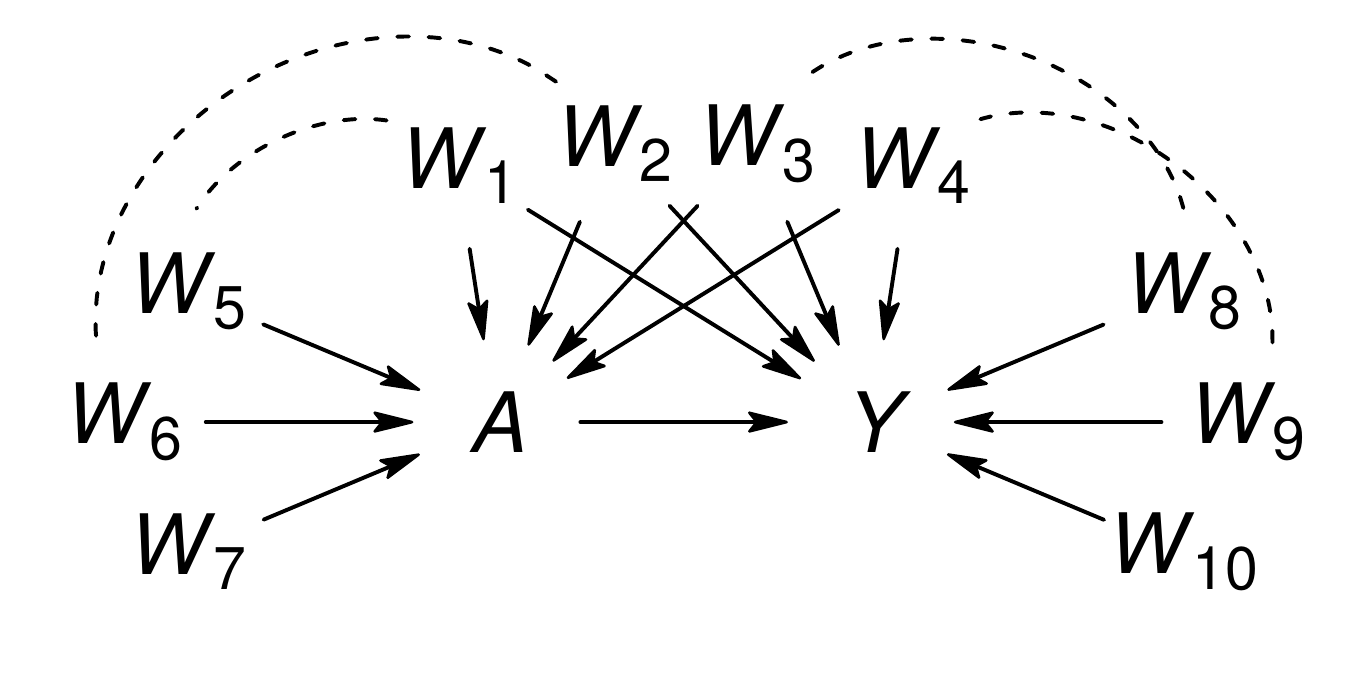}
\caption{Complete data structure for simulation experiments. Dashed arcs without arrowheads connecting variables indicate non-zero entries for the corresponding variables in the covariance matrix of the joint distribution of all $W_i$, $i=1,2,...,10$.}\label{DAG}
\end{figure}

We performed a series of Monte-Carlo simulation experiments based on the simulation structure described in \cite{Setoguchi2008} with modifications so as to allow for missing data.
For $n=2000$ subjects, we independently generated 10 covariates $W_i$ (four confounders, three predictors of the exposure only, and three predictors of the outcome only), a binary exposure variable $A$, and a binary outcome $Y$ (Figure \ref{DAG}). Missing data were introduced into one or two covariates. A number of CART-based approaches were used to estimate propensity scores, before and after the introduction of missing data, and in turn the log odds ratio for the exposure-outcome effect among the treated. For comparison, we also estimated propensity scores in imputed datasets using a correctly specified propensity score model, and using a logistic model with main effects only.
The process was repeated 5000 times for each of eight simulation scenarios that varied primarily by missing data mechanism. All simulations were conducted with R-3.2.2 on a Windows 7 (64-bit) platform \citep{R2016}.

\subsubsection*{Data generation}
Data were generated by sequentially going through the following steps. First, the covariates were generated by sampling from a multivariate normal distribution with zero means and unit variances; correlations were set to zero except for the correlations between $W_1$ and $W_5$, $W_2$ and $W_6$, $W_3$ and $W_8$, and $W_4$ and $W_9$, which were set to $0.2$, $0.9$, $0.2$, and $0.9$, respectively. Second, covariates $W_1$, $W_3$, $W_5$, $W_6$, $W_8$, and $W_9$ were dichotomised, setting any value to 1 if greater than 0 and to 0 otherwise.

Following \cite{Setoguchi2008}, the binary exposure variable $A$ was related to the covariate vector following the propensity score model $\Pr(A=1|W)=\expit\{0.8W_1-0.25W_2+0.6W_3-0.4W_4-0.8W_5-0.5W_6+0.7W_7-0.25W_2^2-0.4W_4^2+0.7W_7^2+0.4W_1 W_3-0.175W_2 W_4+0.3W_3 W_5-0.28W_4 W_6-0.4W_5 W_7+0.4W_1 W_6-0.175W_2 W_3+0.3W_3 W_4-0.2W_4 W_5-0.4W_5 W_6\}$. Realisations $a$ for $A$ were generated by drawing a pseudo-value from the uniform(0,1) distribution and setting $a$ to 1 if this number was less than the true propensity score and to 0 otherwise. Consequently, $A$ can be thought of as a exposure that is generated by a non-linear and non-additive propensity score model. This model assigns approximately half of subjects to the exposure group.

Outcomes were generated following the mechanism described by \cite{Setoguchi2008} with slight modifications to increase the outcome fraction (from approximately 2\% to 20\% or 40\%).
Specifically, the binary outcome, $Y$, was modelled as a  Bernoulli random variable given $A$ and $W$: an independent random number, $\varepsilon_Y$, was drawn from the uniform distribution; $Y$ was set to 1 if this number was less than the inverse logit (expit) of a linear transformation $\eta(A,W)=-1+0.3W_1-0.36W_2-0.73W_3-0.2W_4+0.71W_8-0.19W_9+0.26W_{10}+\gamma A$ of $A$ and $W$ and to 0 otherwise. The true conditional log odds ratio for the exposure-outcome effect was set to $1$ or $-1$ depending on the scenario. The outcome incidence was roughly 40\% for scenarios with $\gamma=1$; and 20\% for scenarios with $\gamma=-1$. The counterfactual outcomes $Y_0$ and $Y_1$ for any subject with realisations $w$ of $W$ and $u$ of $\varepsilon_Y$ are found by computing $I(u<\expit\{\eta(0,w)\})$ and $I(u<\expit\{\eta(1,w)\})$, $I$ denoting the indicator function. With knowledge of the counterfactual outcomes, it can be inferred that with $\gamma=1$, the marginal log odds ratio for the true exposure-outcome effect among the exposed (or treated; ATT) is approximately $0.906$; with $\gamma=-1$, the marginal log odds ratio is approximately $-0.926$ \citep{Hernan2017}. Note that these are different from the conditional causal odds ratios as a result of the non-collapsibility property of the odds ratio.

We considered ignorable missing data mechanisms for introducing missing data.

\textit{MCAR missingness.\qquad}
For all subjects, irrespective of complete data, values of $W_3$ were set to missing with probability $p$, characterising the MCAR mechanism. The missingness probability of the other variables was set to zero.

\textit{MAR missingness.\qquad}
Let $M_3$ be a missing indicator variable that takes the value of one if and only if the value of $W_3$ is missing. Similarly, define $M_4$ to be the missing indicator variable pertaining to $W_4$. Given the full data, $W_3$ and $W_4$ were set to missing independently of one and other and with probability equal to $\Pr(M_3=1|W,A,Y)=p$ and $\Pr(M_4=1|W,A,Y)=\expit\{\alpha_0 + \alpha_1W_1 + \alpha_2A + \alpha_3Y\}$. The missingness probability of the other variables was set to zero.

\subsubsection*{Scenarios}

We evaluated the performance of various CART-based methods in eight scenarios (Table~\ref{Scenarios}). The intercepts $\alpha_0$ in scenarios five through eight were chosen \replaceII{such that the overall rates of missingness (i.e., the fractions of missing data points in the $2000\times12$ datasets) were roughly the same}{so as to yield roughly the same average proportion of missing data points per generated dataset of 24000 data points (2000 records on 10 covariates, one exposure and one outcome variable)}, namely 3\%. The average \replaceII{overall rate of missingness was}{proportion of missing data points and the fraction of incomplete records were} largest in scenario 2 (5\%\addII{ and 60\%, respectively; see Table~\ref{Scenarios}}). In all of the scenarios considered, \replaceII{missingness does not depend on unmeasured or otherwise missing data}{data are `missing at random'} and it is assumed that there is conditional exchangeability given measured covariates (i.e., $(Y_0,Y_1)\CI A|W$).

\begin{table}[h]\centering
\begin{tabular}{cccccccccc}
\toprule
Scenario & $\gamma$ & MCAR/MAR & $p$ & $\alpha_0$ & $\alpha_1$ & $\alpha_2$ & $\alpha_3$ & \addII{PMP} & \addII{PIR}\\
\midrule
1&$\phantom{-}1$&MCAR&0.3&--&--&--&--&\addII{0.03}&\addII{0.30} \\
2&$\phantom{-}1$&MCAR&0.6&--&--&--&--&\addII{0.05}&\addII{0.60} \\
3&$\phantom{-}1$&MAR&0.0&$-0.7$&0.0&0.0&1.5&\addII{0.04}&\addII{0.48} \\
4&$-1$&MAR&0.0&$-1.0$&0.0&0.0&1.5&\addII{0.03}&\addII{0.35} \\
5&$\phantom{-}1$&MAR&0.1&$-1.6$&0.5&0.5&0.5&\addII{0.03}&\addII{0.37} \\
6&$\phantom{-}1$&MAR&0.1&$-2.1$&0.5&0.5&1.5&\addII{0.03}&\addII{0.37} \\
7&$\phantom{-}1$&MAR&0.1&$-2.3$&0.5&1.5&0.5&\addII{0.03}&\addII{0.36} \\
8&$\phantom{-}1$&MAR&0.1&$-2.2$&1.5&0.5&0.5&\addII{0.03}&\addII{0.37} \\
\bottomrule
\end{tabular}
\caption{\deleteII{Simulation parameter values for the various scenarios.}\addII{
Description of scenarios. $\gamma$ equals the conditional log odds ratio for the effect of $A$ on $Y$ given $W$. Given the full data, variables $W_3$ and $W_4$ were set to missing independently of one and other and with probabilities $p$ and $\expit\{\alpha_0+\alpha_1W_1+\alpha_2A+\alpha_3Y\}$, respectively. Abbreviations: MCAR, missing completely at random; MAR, missing at random; PMP, average proportion of missing data points; PIR, average proportion of incomplete records.}}\label{Scenarios}
\end{table}

\delete{Note that CCA estimators tend to be biased in scenarios 5 through 8 as a result of collider stratification \mbox{\citep{Pearl2009}}.} \add{Note that in scenarios 3 trough 8, conditioning on $M$ may break the independence between $A$ and unobserved outcome predictor $\varepsilon_Y$ through what is known as collider stratification (cf. \citeauthor{Pearl2009}, \citeyear{Pearl2009}). One might therefore expect that that discarding incomplete records in these scenarios would result in bias.} In scenarios 3 and 4\add{, however}, covariate missingness $M$ is conditionally independent of exposure status and covariate data given the outcome (i.e., $M\CI (A,W)|Y$). As a result, in these scenarios, the conditional OR for the effect of $A$ on $Y$ given $W$ is equal to the conditional OR given $W$ among the complete cases \citep{Westreich2012}. Bias of \replace{CCA}{complete case} estimators in scenarios 3 and 4 therefore cannot be attributed to collider stratification, despite the presence of an unobserved outcome predictor\delete{ $\varepsilon_Y$}. Instead, it could result from the non-collapsibility of the odds ratio and changes in the covariate distribution brought about by narrowing the focus of inference to the complete cases \citep{Hernan2017}.

\subsubsection*{Estimators}
Bagged CART was based on 100 bootstrap replicates \citep{Lee2010}. 
We imposed complexity constraints on the tree fitting algorithm using the \texttt{rpart} package default control settings.
For boosted CART, we used 20000 iterations, a shrinkage parameter of 0.0005, and an iteration stopping rule based on the mean Kolmogorov-Smirnov test statistic \citep{Lee2010,McCaffrey2004}.

The CART methods were combined with several common approaches to handling missing data: leaving missingness information as is \add{(i.e., subjecting incomplete data directly to the CART algorithm)}; complete case analysis (CCA); and multiple imputation (MI). 
MI was implemented with the \texttt{mice} package (version 2.46.0) using the \texttt{logreg} and \texttt{norm} options to impute missing binary and continuous variables, respectively, and otherwise default settings \citep{mice2011}. Imputation models included, apart from the variable to be imputed, all other variables, including the outcome, as untransformed main effects only. Propensity score analysis was performed within imputed datasets using the respective sets of estimated propensity scores \citep{PenningdeVries2016} and results were combined using \citeauthor{Rubin1987}'s (\citeyear{Rubin1987}) rules.

In addition to using CART, as stated, we also estimated propensity scores in imputed datasets using a correctly specified propensity score model (LRc), and using a logistic model with main effects only (LRm).

Within each (multiply imputed) dataset, the ATT was estimated from a logistic model with robust variance estimation using the \texttt{survey} package \citep[version 3.31]{survey2014}. We used both propensity score matching and inverse probability weighting. Matching was performed using a greedy 1:1 nearest neighbour algorithm, matching exposed ($A=1$) to unexposed individuals ($A=0$) \citep{Austin2011intro}. For any given (imputed) dataset, matching was performed on the logit of the propensity score, using a calliper distance of 20\% of the standard deviation of the logit propensity score estimates \citep{Austin2011}. With the ATT as the estimand, IPW weights were defined as 1 for exposed subjects and $\mathrm{PS/(1-PS)}$ for unexposed subjects (PS denoting the estimated propensity score). To avoid undefined weights (1/0) or logit propensity scores ($\mathrm{logit}(0)$), we placed bounds on the estimated propensity scores, truncating all estimates less than 0.001 to 0.001 and setting estimates greater than 0.999 to 0.999.
MI-based estimates were pooled using Rubin's rules to yield for each original dataset a single effect estimate, standard error estimate, and 90\% confidence interval (90\%CI).

\subsubsection*{Performance metrics}
We evaluated the performance of the various methods through several measures: bias, estimated by the mean deviation of the estimated from the true marginal exposure-outcome effect on the log scale; empirical standard error; mean estimated standard error; mean squared error (MSE); and 90\%CI coverage, estimated by the percentage of the 5000 data sets in which the 90\%CI included the true exposure-outcome effect. Based on 5000 simulation runs, the Monte Carlo standard error for the true coverage probability of 0.90 is $\surd(0.90(1-0.90)/5000)\approx 0.0042$, implying that the estimated coverage probability is expected to lie with 95\% probability between 0.893 and 0.907. Empirical coverage rates outside this interval provide evidence against the true coverage probabilities being equivalent to the nominal level of 0.90. The primary interest, however, was to gauge the effect of missing data on the various effect estimators. Therefore, we also compared, for each scenario, the effect estimates before and after the introduction of missing data. 

\subsection{Results}

In this section, we present \add{(Table~\ref{IPW_res})} and describe  the results for IPW-based estimators only. Trends for estimators based on propensity score matching are similar and the results are presented in full in the Supplementary Material.

\subsubsection*{Bias} Before the introduction of missing data, baCART and bCART showed small to no bias (\replace{\textless2\% of the true effect}{with absolute values ranging from 0.000 to 0.011} on the log odds ratio scale). MI+LRc performed generally well and to a similar extent as MI+baCART and MI+bCART. MI+LRm consistently underestimated the true effect when inference was based on IPW; this trend was weaker for inference based on propensity score matching (Supplementary Table 1). Among all CART-based missing data approaches considered, multiple imputation yielded the least biased estimators overall\add{ (with a maximum absolute value of bias of 0.026 versus 0.221 and 0.138 for CART-only and CCA estimators, respectively)}, whereas bCART deviated on average the most from the true effect after the introduction of missingness.

As expected, baCART and bCART were biased \add{(with $-$0.029 and $-$0.039, respectively, for scenario 1 and $-$0.064 and $-$0.072 for scenario 2) }under MCAR in the direction of confounding by $W_3$, whereas CCA and MI produced exposure-outcome effect estimates that were on average very close to the true effect.
In scenarios 3 and 4, where missingness was outcome-dependent, bCART was biased toward the null after the introduction of missingness\add{ (with bias estimates of $-$0.117 and 0.064 for scenario 3 and 4, respectively, where the causal log odds ratios were approximately 0.906 and $-$0.926)}; baCART was downwardly biased\add{ in both scenarios (with bias estimates of $-$0.088 and $-$0.112)}. Estimators based on CCA or MI with CART were considerably less biased.
In scenarios 5 through 8, CCA estimators systematically underestimated the true effect, particularly when the effect of the exposure or the outcome on the missingness probability was large\add{ (scenarios 6 and 7, where bias estimates ranged from $-$0.116 to $-$0.138)}. \add{In these scenarios (5 through 8),} baCART produced estimates that were on average close to the true effect, except in scenario 6, where the effect was clearly underestimated \add{(estimated bias $-$0.050)}. bCART resulted in estimates that deviated in the same direction and to a similar or greater extent from the true effect as compared with CCA estimators. Again, MI with CART resulted in estimates that were on average close to the true effect. Increasing the effect of covariate $W_1$ on the missingness probability (scenario 8 versus 5) had no evident impact on the results of any of the estimators.

\subsubsection*{Other performance}
As expected, discarding incomplete records (CCA) resulted in relatively large empirical standard errors. Interestingly, MI+LRc had the largest empirical standard error in most scenarios, probably as a consequence of the complexity of the fitted propensity score models. In comparing empirical and mean estimated standard errors, note that multiple imputation produced generally conservative estimates of the standard error. This is consistent with previous observations \citep{Buuren2012}. Among the CART-based estimators, the MSE was largest for CCA in nearly all scenarios. MI estimators had consistently small MSE. Overall, the best performance in terms of MSE was attained by MI estimators, followed by baCART and bCART.
Multiple imputation with CART resulted in empirical coverage rates close to or slightly higher than the nominal 90\% and those of the other estimators.

\add{
\subsection{Additional simulation experiment}
To investigate the estimator performances in a simpler setting, we repeated the simulation experiment of scenario 2 with the squared and interaction terms removed from the exposure allocation model of the data generating mechanism. The results, presented in Supplementary Table 2, indicate generally the same trends as previously noted. Of note, in the absence of missing data, inverse weighting based on CART showed noticeably more bias than in scenarios 1 through 8. This is probably related to CART's inherent limited ability to model smooth functions. Multiply imputing missing data followed by CART yielded approximately the same extent of bias. However, this bias appears to be partially cancelled out by the bias introduced by CART's automatic handling of missing data to the extent that CART alone performed better with than without missing data. Nonetheless, relative to the extent of bias of the respective CART algorithm before the introduction of missing data, multiple imputation with CART outperformed both CCA with CART and CART applied directly to incomplete data in terms of bias.
}

\begin{landscape}\captionsetup{width=\linewidth}
\begin{longtable}{llld{1.3}d{1.3}d{1.3}d{1.3}d{1.3}d{1.3}d{1.3}d{1.3}}
\toprule
&Missing&&\multicolumn{8}{c}{Scenario}\\
\cline{4-11}
Metric&data&Method&\multicolumn{1}{c}{1}&\multicolumn{1}{c}{2}&\multicolumn{1}{c}{3}&\multicolumn{1}{c}{4}&\multicolumn{1}{c}{5}&\multicolumn{1}{c}{6}&\multicolumn{1}{c}{7}&\multicolumn{1}{c}{8} \\
\midrule\endhead
\midrule\multicolumn{11}{r}{Continued on next page.}
\endfoot
\caption{Performance metrics of inverse probability weighting estimators in 5000 simulated datasets with and without missing data. Abbreviations: SE, standard error; MSE, mean squared error; 90\%CI, 90\% confidence interval; CART, classification and regression trees; baCART, bootstrap aggregated CART; bCART, boosted CART; CCA, complete case analysis; MI, multiple imputation; LRc, logistic regression with correctly specified model; LRm, logistic regression with main effects only.}\label{IPW_res}\endlastfoot
Bias & Without & baCART & 0.009 & 0.011 & 0.009 & 0.011 & 0.007 & 0.007 & 0.008 & 0.010 \\
 &  & bCART & -0.001 & 0.002 & -0.000 & 0.004 & -0.001 & -0.000 & -0.001 & 0.001 \\
 & With & baCART & -0.029 & -0.064 & -0.088 & -0.112 & -0.011 & -0.050 & 0.010 & -0.004 \\
 &  & bCART & -0.037 & -0.072 & -0.117 & 0.064 & -0.057 & -0.221 & -0.123 & -0.053 \\
 &  & CCA+baCART & 0.001 & 0.001 & 0.016 & -0.023 & -0.040 & -0.138 & -0.116 & -0.032 \\
 &  & CCA+bCART & 0.000 & 0.006 & 0.022 & -0.010 & -0.046 & -0.136 & -0.129 & -0.035 \\
 &  & MI+baCART & 0.001 & 0.002 & 0.025 & 0.018 & 0.020 & 0.016 & 0.022 & 0.026 \\
 &  & MI+bCART & -0.008 & -0.011 & -0.024 & -0.025 & -0.010 & -0.023 & -0.008 & -0.007 \\
 &  & MI+LRc & 0.002 & -0.007 & -0.028 & -0.029 & -0.007 & -0.025 & -0.004 & -0.002 \\
 &  & MI+LRm & -0.099 & -0.094 & -0.072 & -0.075 & -0.087 & -0.083 & -0.089 & -0.082 
\\ &&&&&&&&&& \\
Empirical & Without & baCART & 0.116 & 0.114 & 0.116 & 0.129 & 0.115 & 0.115 & 0.114 & 0.116 \\
SE &  & bCART & 0.134 & 0.133 & 0.136 & 0.147 & 0.135 & 0.133 & 0.133 & 0.135 \\
 & With & baCART & 0.118 & 0.121 & 0.126 & 0.136 & 0.119 & 0.121 & 0.119 & 0.119 \\
 &  & bCART & 0.132 & 0.128 & 0.125 & 0.137 & 0.131 & 0.127 & 0.149 & 0.131 \\
 &  & CCA+baCART & 0.141 & 0.186 & 0.189 & 0.199 & 0.147 & 0.156 & 0.143 & 0.148 \\
 &  & CCA+bCART & 0.158 & 0.202 & 0.211 & 0.221 & 0.165 & 0.172 & 0.154 & 0.163 \\
 &  & MI+baCART & 0.116 & 0.116 & 0.115 & 0.129 & 0.114 & 0.115 & 0.113 & 0.116 \\
 &  & MI+bCART & 0.132 & 0.130 & 0.129 & 0.140 & 0.128 & 0.126 & 0.126 & 0.128 \\
 &  & MI+LRc & 0.216 & 0.205 & 0.202 & 0.218 & 0.210 & 0.211 & 0.203 & 0.214 \\
 &  & MI+LRm & 0.125 & 0.123 & 0.125 & 0.138 & 0.122 & 0.121 & 0.124 & 0.123 
\\ &&&&&&&&&& \\
Mean & Without & baCART & 0.114 & 0.114 & 0.114 & 0.128 & 0.114 & 0.114 & 0.114 & 0.114 \\
$\widehat{\text{SE}}$ &  & bCART & 0.136 & 0.136 & 0.136 & 0.149 & 0.136 & 0.136 & 0.136 & 0.136 \\
 & With & baCART & 0.115 & 0.119 & 0.118 & 0.129 & 0.117 & 0.117 & 0.119 & 0.117 \\
 &  & bCART & 0.134 & 0.132 & 0.131 & 0.144 & 0.134 & 0.135 & 0.153 & 0.135 \\
 &  & CCA+baCART & 0.139 & 0.189 & 0.190 & 0.199 & 0.148 & 0.156 & 0.145 & 0.148 \\
 &  & CCA+bCART & 0.160 & 0.204 & 0.211 & 0.221 & 0.166 & 0.174 & 0.160 & 0.163 \\
 &  & MI+baCART & 0.116 & 0.120 & 0.115 & 0.130 & 0.116 & 0.116 & 0.115 & 0.116 \\
 &  & MI+bCART & 0.140 & 0.143 & 0.137 & 0.150 & 0.138 & 0.138 & 0.137 & 0.138 \\
 &  & MI+LRc & 0.196 & 0.198 & 0.187 & 0.201 & 0.189 & 0.191 & 0.185 & 0.191 \\
 &  & MI+LRm & 0.131 & 0.135 & 0.128 & 0.142 & 0.128 & 0.128 & 0.128 & 0.128 
\\ &&&&&&&&&& \\
MSE & Without & baCART & 0.014 & 0.013 & 0.013 & 0.017 & 0.013 & 0.013 & 0.013 & 0.014 \\
 &  & bCART & 0.018 & 0.018 & 0.019 & 0.022 & 0.018 & 0.018 & 0.018 & 0.018 \\
 & With & baCART & 0.015 & 0.019 & 0.024 & 0.031 & 0.014 & 0.017 & 0.014 & 0.014 \\
 &  & bCART & 0.019 & 0.022 & 0.029 & 0.023 & 0.020 & 0.065 & 0.037 & 0.020 \\
 &  & CCA+baCART & 0.020 & 0.034 & 0.036 & 0.040 & 0.023 & 0.043 & 0.034 & 0.023 \\
 &  & CCA+bCART & 0.025 & 0.041 & 0.045 & 0.049 & 0.029 & 0.048 & 0.040 & 0.028 \\
 &  & MI+baCART & 0.013 & 0.013 & 0.014 & 0.017 & 0.013 & 0.013 & 0.013 & 0.014 \\
 &  & MI+bCART & 0.018 & 0.017 & 0.017 & 0.020 & 0.016 & 0.016 & 0.016 & 0.016 \\
 &  & MI+LRc & 0.047 & 0.042 & 0.042 & 0.048 & 0.044 & 0.045 & 0.041 & 0.046 \\
 &  & MI+LRm & 0.025 & 0.024 & 0.021 & 0.025 & 0.023 & 0.021 & 0.023 & 0.022 
\\ &&&&&&&&&& \\
Empirical & Without & baCART & 0.896 & 0.901 & 0.895 & 0.897 & 0.892 & 0.899 & 0.898 & 0.890 \\
90\%CI &  & bCART & 0.907 & 0.909 & 0.903 & 0.904 & 0.904 & 0.909 & 0.909 & 0.907 \\
coverage & With & baCART & 0.881 & 0.847 & 0.784 & 0.753 & 0.890 & 0.854 & 0.898 & 0.893 \\
 &  & bCART & 0.897 & 0.861 & 0.772 & 0.886 & 0.878 & 0.497 & 0.797 & 0.880 \\
 &  & CCA+baCART & 0.893 & 0.905 & 0.894 & 0.901 & 0.888 & 0.760 & 0.792 & 0.884 \\
 &  & CCA+bCART & 0.906 & 0.905 & 0.902 & 0.904 & 0.890 & 0.798 & 0.804 & 0.886 \\
 &  & MI+baCART & 0.904 & 0.914 & 0.894 & 0.897 & 0.895 & 0.900 & 0.903 & 0.896 \\
 &  & MI+bCART & 0.922 & 0.931 & 0.915 & 0.919 & 0.918 & 0.928 & 0.926 & 0.923 \\
 &  & MI+LRc & 0.911 & 0.920 & 0.909 & 0.906 & 0.908 & 0.921 & 0.919 & 0.906 \\
 &  & MI+LRm & 0.815 & 0.841 & 0.858 & 0.865 & 0.831 & 0.848 & 0.827 & 0.841 
\\\bottomrule
\end{longtable}
\end{landscape}

\section{Case study}\label{CaseStudy}

In this section, we illustrate the application of the CART-based estimators to an empirical dataset, constructed to assess the association between annual influenza vaccination and mortality risk among elderly \citep{Groenwold2009}. The dataset comprises 44418 complete records on vaccination status, mortality during the influenza epidemic period and potential confounders (age, sex, health status and prior health care and medication use). Among the 32388 vaccinated individuals 266 died, whereas 113 out of 12030 nonvaccinated individuals died (crude odds ratio 0.87, 90\%CI 0.73--1.05). To control for measured confounders, propensity scores were estimated via bCART and a pseudopopulation was constructed using IPW such as to preserve the covariate distribution of the vaccination group. This resulted in an odds ratio of 0.60 (90\%CI 0.49--0.73) for the marginal effect of vaccination on mortality risk among the vaccinated. Substituting bCART with baCART yielded an odds ratio of 0.65 (90\% 0.53--0.81). As expected, introducing MCAR missingness into a confounder by setting a random 50\% of subjects' number of prior general practitioner (GP) visits to missing, resulted in odds ratio estimates that were closer to the crude effect.
Setting the number of GP visits to missing with probability 0.5 for all subjects who died and zero otherwise, resulted in estimates substantially closer to the null for bCART and away from the null for baCART. Thus, as in our simulations, outcome-dependent MAR missingness resulted in apparent attenuation of the exposure-outcome effect as estimated by bCART. Table \ref{Res_vaccination} shows the results also for the complete case and multiple imputation equivalents of baCART and bCART as well as for IPW based on propensity score estimation using main effects logistic regression and with weights truncated to the interval from the 0th to the 97.5th percentile. To better handle potential violations of standard imputation model assumptions, we used a nonparametric multiple imputation strategy (option \texttt{cart} rather than \texttt{norm} in the \texttt{mice} package) to estimate the effect of vaccination on mortality risk \citep{Doove2014}. Interestingly, in the MAR setting, bCART and baCART yielded the two most extreme estimates for the effect of vaccination on mortality risk among the elderly.

\begin{table}\centering
\setlength{\tabcolsep}{4pt}
\begin{tabular}{lccccc}
\toprule
&\multicolumn{5}{c}{Missingness} \\\cline{2-6}
&None&&MCAR&&MAR \\\cline{2-2}\cline{4-4}\cline{6-6}
Method&OR (90\%CI)&&OR (90\%CI)&&OR (90\%CI)\\
\midrule
baCART&0.65 (0.53--0.81)&&0.69 (0.56--0.85)&&0.53 (0.44--0.66)\\
bCART&0.60 (0.49--0.73)&&0.63 (0.51--0.77)&&0.79 (0.63--0.98)\\
CCA+baCART&--&&0.55 (0.41--0.73)&&0.62 (0.46--0.84)\\
CCA+bCART&--&&0.50 (0.37--0.66)&&0.56 (0.42--0.75)\\
MI+baCART&--&&0.60 (0.47--0.75)&&0.70 (0.55--0.89)\\
MI+bCART&--&&0.58 (0.47--0.72)&&0.63 (0.51--0.78)\\
LRm\textsuperscript{\textdagger}&0.59 (0.49--0.71)&&0.62 (0.51--0.76)&&0.70 (0.57--0.86)
\\\bottomrule
\end{tabular}
\caption{Estimated effects of vaccination on mortality risk among the elderly in dataset with no missing data, MCAR missingness or outcome-dependent MAR missingness. Estimates are adjusted for age, sex, health status and prior health care and medication use. Abbreviations: MCAR, missing completely at random; MAR, missing at random; OR, odds ratio; 90\%CI, 90\% confidence interval; CART, classification and regression trees, baCART, bootstrap aggregated CART; bCART, boosted CART; CCA, complete case analysis; MI, multiple imputation; LRm, main effects logistic regression. \textsuperscript{\textdagger}In case of (MCAR or MAR) missingness, MI was implemented before LRm.}\label{Res_vaccination}
\end{table}

\section{Discussion}\label{Discussion}

In this paper we examined the workings of CART based propensity score estimators in scenarios with missing covariate data. Although the CART has been described as a promising approach to automatically handle missing covariate data when developing a propensity score \citep{Setoguchi2008,Lee2010}, there has been little discussion on the performance of these methods. Through analysis and simulations we showed that the application of CART for propensity score estimation can yield serious bias in estimates of exposure-outcome relations. We showed that this problem not only pertains to the situation of MAR but critically also to the situations with MCAR, which are often considered harmless when bias is concerned resulting only in larger variance of the estimator of exposure-outcome relations.

An attractive property of CART-based methods relative to standard logistic regression procedures, is perhaps not having to discard incomplete records. Indeed, in our simulations, discarding incomplete records resulted in the largest empirical standard errors. 
Alternatively, multiple imputation may be used to replace missing values under MCAR or MAR prior to propensity score estimation. This approach was shown to work well in our simulations. One criticism of multiple imputation in its parametric form is that it makes possibly erroneous distributional assumptions. In particular, the standard multiple imputation algorithms do not properly capture nonlinear relations like interaction effects \citep{Cham2016}. Multiple imputation algorithms that use nonparametric methods have been developed. For example, \cite{Doove2014}, following \cite{Burgette2010}, proposed CART to be incorporated as imputation method in the multiple imputation by chained equations framework. As with parametric multiple imputation, the algorithm is designed to account for the inherent variability in the data.
However, while the approach of \cite{Doove2014} seems promising, there is still room for improvement. Particularly, the algorithm does not explicitly account for uncertainty about the (implicit) CART trees' model parameters. 
To address this, \cite{Shah2014} proposed a promising algorithm in which random forest CART is embedded in the multiple imputation by chained equations framework and imputation models are fitted to bootstrap samples.  
An implementation is available via the R package \texttt{CALIBERrfimpute} \citep{rfimpute}.

In interpreting our findings, it is important to note that we considered only a small number of scenarios. We assumed throughout that data were MCAR or MAR and that there was no unmeasured confounding (conditional exchangeability given measured confounders). As noted, there are situations conceivable in which it is not problematic to estimate the generalised propensity score. If the missingness information conveys information about a strong unmeasured confounder, estimating the generalised propensity score may allow for partial control of unmeasured confounding. On the other hand, adjusting for missingness information, e.g., through the generalised propensity score (estimated by some CART algorithms), may be problematic particularly when it is a strong proxy for the outcome, an intermediate, or a common effect of the exposure and outcome.

\add{Our arguments for caution when using CART to estimate propensity scores in the presence of missing data are in line with the recommendation to incorporate information on the outcome in imputing missing covariate data \citep{PenningdeVries2016,Leyrat2017,Moons2006}. Since propensity score estimation is typically done without any information on the outcome \citep{Rubin2008}, any missing data imputation (e.g., with a surrogate) that is inherent to the propensity score estimation procedure will likely fail. An important feature of the propensity score matching or weighting methodology is that, in the absence of missing data, it need not make distributional assumptions about the outcome in relation to the exposure and covariates in constructing a matched or weighted dataset. In the presence of missing covariate data, omitting information on the outcome in imputing missing covariate data, however, imposes a structure on the data that likely contrasts with the true data distribution and the analysis model. This is similar to the idea of models being ``uncongenial'' in the sense of \cite{Meng1994}.
The current study also relates to the literature on the missing indicator method, given its resemblance with the approach to handling missing data taken by the boosted CART algorithm. Like the automatic handing of missing data by the boosted CART algorithm, the missing indicator method typically results in bias \citep{Groenwold2012}.
}

It has been suggested to perform balance diagnostics on the matched or weighted study sample at hand \citep{Austin2011intro}. If systematic differences persist between exposure groups following matching or weighting, this may be an indication that the propensity score estimation algorithm requires modification \citep{Austin2015}. In the context of CART, one may assign greater weight to subjects at a certain covariate level in evaluating exposure homogeneity at any given node. We did not adopt an iterative approach to propensity score estimation and balance diagnostics in our simulation studies for several reasons. First, doing so would increase the computational burden of the simulations. Second, whereas CART facilitates the estimation of propensity scores that balance the entire covariate joint distribution across exposure groups, standard balance diagnostics procedures typically ignore the complex relationship between exposure and covariates. For example, when using the standardised mean difference, it is typically assumed that all variables that need to be balanced with respect to the mean are identified and included in the set over which a summary (e.g., weighted mean or maximum) standardised mean difference is calculated. The utility of the metric may be poor if important variables (e.g., higher order moments) are omitted. Other balance metrics, such as the Kolomogorov-Smirnov metric, L\'{e}vy distance, and overlapping coefficient \citep{Belitser2011,Franklin2014,Ali2015} often fail to reflect the extent of imbalance with respect to the entire covariate joint distribution. In addition, what constitutes good balance ultimately depends on the outcome model too. Substantial imbalance may be acceptable for covariates that are weakly predictive of the outcome, while small departures from perfect balance may be problematic for covariates that are strongly predictive of the outcome.

We emphasise that our simulations were not designed to compare CART versus logistic regression as means to estimate propensity scores. Main effects logistic regression here and in previous studies demonstrated a robust performance against model misspecification in terms of bias when inference was based on propensity score matching \citep{Setoguchi2008}. This is likely attributable to the set-up of the simulations. The outcome model included homogeneous exposure-outcome effects and main effects only. Since between-exposure-group imbalances with respect to interaction terms or higher order moments of covariates need not accompany systematic differences in outcomes, it is not surprising that propensity score matching based on main effects logistic regression may perform roughly the same in terms of bias as propensity score matching based on logistic regression with correct model specification. Further studies comparing CART versus main effects logistic regression may well demonstrate more clearly the advantageous properties of CART in settings with both complex propensity score and complex outcome models.

In summary, we compared various approaches to handling missing data in estimating propensity scores via CART. While the use of machine learning in estimating propensity scores seems promising for handling complex full data structures, it unlikely represents a suitable substitute for well-established methods, such as multiple imputation, to deal with missing data.

\section*{Appendices}

\subsection*{Appendix A}
For realisations $w$ of $W$ and $a$ of $A$, let
$$\varphi(w,a)=\frac{\varphi^\ast(w,a)}{\mathbb{E}[\varphi^\ast(W,A)|A=a]},\qquad \varphi^\ast(w,a)=I(a=1)+I(a=0)\frac{e(w)}{1-e(w)}.$$
We show in this subsection that weighting by $\varphi$ yields independence between covariate(s) $W$ and $A$; that is, for all $w$, $$\varphi(w,0)\Pr(W=w|A=0)=\varphi(w,1)\Pr(W=w|A=1).$$ Also, conditional exchangeability given $W$ implies exchangeability following weighting by $\varphi$; that is, if $(Y_0,Y_1)\CI A|W=w$ for all $w$, then \begin{align*}\sum_w\varphi(w,0)\Pr(Y_0&=y_0,Y_1=y_1,W=w|A=0)\\&=\sum_w\varphi(w,1)\Pr(Y_0=y_0,Y_1=y_1,W=w|A=1)\end{align*} for all $y_0,y_1$.

We begin by considering $\mathbb{E}[\varphi^\ast(W,A)|A=a]$. It is easy to see that $\mathbb{E}[\varphi^\ast(W,A)|A=1]=1$. For $a=0$, we have
\begin{align*}
\mathbb{E}[\varphi^\ast(W,A)|A=0]
	&= \mathbb{E}\bigg[\frac{e(W)}{1-e(W)}\Big|A=0\bigg] \\
&= \mathbb{E}\bigg[\frac{\Pr(A=1|W)}{\Pr(A=0|W)}\Big|A=0\bigg] \\
&= \sum_{w}\frac{\Pr(A=1|W=w)}{\Pr(A=0|W=w)}\Pr(W=w|A=0) \\
&= \sum_{w}\frac{\Pr(W=w|A=1)\Pr(A=1)/\Pr(W=w)}{\Pr(W=w|A=0)\Pr(A=0)/\Pr(W=w)}\Pr(W=w|A=0) \\
&= \frac{1}{\Pr(A=0)}\sum_w\Pr(W=w|A=1)\Pr(A=1) \\
&= \frac{\Pr(A=1)}{\Pr(A=0)}
\end{align*}

Since $\varphi(w,1)=1$, to prove the first statement it suffices to show that $\varphi(w,0)\Pr(W=w|A=0)=\Pr(W=w|A=1)$ for all $w$. Now,
\begin{align*}
\varphi(w,0)\Pr(W&=w|A=0)
\\&= \frac{e(w)}{1-e(w)}\frac{\Pr(A=0)}{\Pr(A=1)}\Pr(W=w|A=0) \\
&= \frac{\Pr(A=1|W=w)}{\Pr(A=0|W=w)}\frac{\Pr(A=0)}{\Pr(A=1)}\Pr(W=w|A=0) \\
&= \frac{\Pr(W=w|A=1)\Pr(A=1)}{\Pr(W=w|A=0)\Pr(A=0)}\frac{\Pr(A=0)}{\Pr(A=1)}\Pr(W=w|A=0) \\
&= \Pr(W=w|A=1),
\end{align*}
for all $w$, as desired.

To complete this proof, observe that
\begin{align*}
\sum_w\varphi(w,0)&\Pr(Y_0=y_0,Y_1=y_1,W=w|A=0) \\
	&= \sum_w\frac{\Pr(A=1|W=w)}{\Pr(A=0|W=w)}\frac{\Pr(A=0)}{\Pr(A=1)}\Pr(Y_0=y_0,Y_1=y_1,W=w|A=0)\\
&= \sum_w\frac{\Pr(W=w|A=1)\Pr(A=1)/\Pr(W=w)}{\Pr(W=w|A=0)\Pr(A=0)/\Pr(W=w)}\frac{\Pr(A=0)}{\Pr(A=1)}\\&\qquad\times\Pr(Y_0=y_0,Y_1=y_1|W=w,A=0)\Pr(W=w|A=0) \\
&= \sum_w\Pr(W=w|A=1)\Pr(Y_0=y_0,Y_1=y_1|W=w,A=0)
\end{align*}
for all $y_0,y_1$.
Under conditional exchangeability given $W$, i.e., $(Y_0,Y_1)\CI A|W$, we have $\Pr(Y_0=y_0,Y_1=y_1|W=w,A=0)=\Pr(Y_0=y_0,Y_1=y_1|W=w,A=1)$ for all $w$. Hence, $\sum_w\varphi(w,0)\Pr(Y_0=y_0,Y_1=y_1,W=w|A=0)$ becomes $\sum_w\Pr(W=w|A=1)\Pr(Y_0=y_0,Y_1=y_1|W=w,A=1)$, which is equal to $\Pr(Y_0=y_0,Y_1=y_1|A=1)$. Since $\varphi(w,1)=1$, we also have that $\sum_w\varphi(w,1)\Pr(Y_0=y_0,Y_1=y_1,W=w|A=1)=\Pr(Y_0=y_0,Y_1=y_1|A=1)$ for all $y$, which completes this proof.

\subsection*{Appendix B}
In this subsection, we give an example of a simple setting where $(Y_0,Y_1)\CI A|e(W)$ and $W\CI A|e^\ast(V)$ hold, yet $(Y_0,Y_1)\nCI A|e^\ast(V)$.

Let $W$, $A$ and $Y$ be binary mutually independent random variables and suppose that covariate missingness is MAR dependent on $Y$. Specifically, let $\Pr(W=1)=0.5$ and $\Pr(A=1|W=w)=\Pr(A=1)=0.5$ for all $w$. Further, define $Y=I(\varepsilon_Y<(1+A)/10)$, where $\varepsilon_Y\sim\mathcal{U}(0,1)$ such that $\varepsilon_Y\CI (A,W)$. Thus, there is conditional exchangeability given $W$, so that $\Pr(Y=1|A=a,W=w)=\Pr(Y_a=1|A=a,W=w)=(1+a)/10$ for all $a,w$. \citeauthor{Rosenbaum1983} (\citeyear{Rosenbaum1983}, Theorems 1 and 3) and \replace{Appendix~B}{Appendix~A} establish conditional exchangeability given $e(W)$ and exchangeability following inverse probability weighting with weights defined on the basis of $e(W)$. Now, let $\Pr(R=0|W,A,Y,\varepsilon_Y)=0.1+0.5Y$. It is easily verified that $W\CI A|e^\ast(V)$. However, $e^\ast(V)=4/7$ if and only if $V=\ast$ or, equivalently, $R=0$. Since $R\CI \varepsilon_Y|(A,Y)$ and $R\CI A|Y$, for any $u\in (0,1)$, we therefore have
\begin{align*}
\Pr(\varepsilon_Y\le u|A&=a,e^\ast(V)=4/7) \\ &= \Pr(\varepsilon_Y\le u|A=a,R=0) \\
	&= \sum_y\Pr(\varepsilon_Y\le u|A=a,Y=y)\Pr(Y=y|A=a,R=0) \\
&= \sum_y\bigg\{\Pr(\varepsilon_Y\le u|A=a,Y=y) \\
	&\qquad\times \frac{\Pr(R=0|Y=y)\Pr(Y=y|A=a)}{\sum_{y'}\Pr(R=0|Y=y')\Pr(Y=y'|A=a)}\bigg\} \\
	&= \sum_y \Pr(\varepsilon_Y\le u|A=a,Y=y) \frac{(1+5y)[(1+a)y+(9-a)(1-y)]}{15+5a},
\end{align*}
where\
\begin{align*}
\Pr(\varepsilon_Y\le u|A=a,Y=y)
	&= \frac{\Pr(Y=y|A=a,\varepsilon_Y\le u)\Pr(\varepsilon_Y\le u)}{\Pr(Y=y|A=a)} \\
&= \frac{q(y,u,a)u}{(1+a)y/10 + (9-a)(1-y)/10},
\end{align*}
with $q(y,u,a)=1-y+(-1)^{1-y}\mathrm{min}\{(1+a)/10,u\}/u$. In particular, $\Pr(\varepsilon_Y\le 0.5|A=a,e^\ast(V)=4/7)$ equals $2/3$ if $a=0$ and $3/4$ if $a=1$. Hence, $\varepsilon_Y\nCI A|e^\ast(V)$ and, given the definitions of $Y$, $Y_0$ and $Y_1$, we have $(Y_0,Y_1)\nCI A|e^\ast(V)$.

\subsection*{Appendix C}

This subsection details an example where $(Y_0,Y_1)\CI A|e^\ast(V)$, yet $(Y_0,Y_1)\nCI A|e(W)$.

Suppose that $W$ and that $A$ and $Y$ are all binary random variables. Further, let $(A,R)$ be marginally independent of $W$, let $A$ conditionally depend on $R$ given $W$, and let $Y$ conditionally depend on $A$ and $R$ given $W$. Specifically, let $\Pr(W=1)=0.5$, $\Pr(R=0|W)=0.1$, $\Pr(A=1|R,W)=2(1+R)/10$, and $Y=I(\varepsilon_Y<2(1+2R)/20)$, where $\varepsilon_Y\CI (W,R,A)$. To see that $(Y_0,Y_1)\CI A|e^\ast(V)$, first note that
\begin{align*}
e(w) &= \Pr(A=1|W=w) \\
	&= \Pr(A=1|W=w,R=0)\Pr(R=0|W=w) \\&\qquad+\Pr(A=1|W=w,R=1)\Pr(R=1|W=w) \\
&= 0.38,
\end{align*}
for $w=0,1$, and that $e^\ast(v)$ equals $\Pr(A=1|R=0)=0.20$ if $v=\ast$ and $\Pr(A=1|R=1)=0.40$ otherwise. Now,
\begin{align*}
\Pr(Y_0=1|A=a,e^\ast(V)=0.20) &= \Pr(Y_0=1|A=a,e^\ast(V)=0.20) \\
	&= \Pr(Y_0=1|A=a,R=0) \\
&= \Pr(\varepsilon_Y<2(1+2R)/20|A=a,R=0) \\
&= \Pr(\varepsilon_Y<0.10|A=a,R=0) \\
&= \Pr(\varepsilon_Y<0.10) = 0.10
\end{align*}
for $a=0,1$. Also, $\Pr(Y_0=1|A,e^\ast(V)=0.40) =0.10$. Thus, $Y_0\CI A|e^\ast(V)$. Similarly, it can be shown that $(Y_0,Y_1)\CI A|e^\ast(V)$.  Next, observe that \begin{align*}
\Pr(Y_0=1|A=a,e(W)=0.38) &= \Pr(Y_0=1|A=a) \\
	&= \Pr(Y_0=1|A=a,R=0)\Pr(R=0|A=a) \\&\qquad+ \Pr(Y_0=1|A=a,R=1)\Pr(R=1|A=a) \\
&= 0.10\frac{\Pr(A=a|R=0)\Pr(R=0)}{\Pr(A=a)} \\&\qquad+ 0.30\frac{\Pr(A=a|R=1)\Pr(R=1)}{\Pr(A=a)} \\
&= \frac{0.20^a0.80^{1-a}0.01+0.40^a0.60^{1-a}0.27}{0.38^a0.62^{1-a}},
\end{align*}
which is not invariant to changes in $a=0,1$. Hence, $(Y_0,Y_1)\nCI A|e(W)$.

\begin{landscape}\captionsetup{width=\linewidth}
\section*{Supplementary Material}

\begin{longtable}{llld{1.3}d{1.3}d{1.3}d{1.3}d{1.3}d{1.3}d{1.3}d{1.3}}
\toprule
&Missing&&\multicolumn{8}{c}{Scenario}\\
\cline{4-11}
Metric&data&Method&\multicolumn{1}{c}{1}&\multicolumn{1}{c}{2}&\multicolumn{1}{c}{3}&\multicolumn{1}{c}{4}&\multicolumn{1}{c}{5}&\multicolumn{1}{c}{6}&\multicolumn{1}{c}{7}&\multicolumn{1}{c}{8} \\
\midrule\endhead
\midrule\multicolumn{11}{r}{Continued on next page.}
\endfoot
\caption*{Supplementary Table 1: Performance metrics of propensity score matching estimators in 5000 simulated datasets with and without missing data. Abbreviations: SE, standard error; MSE, mean squared error; 90\%CI, 90\% confidence interval; CART, classification and regression trees; baCART, bootstrap aggregated CART; bCART, boosted CART; CCA, complete case analysis; MI, multiple imputation; LRc, logistic regression with correctly specified model; LRm, logistic regression with main effects only.}\endlastfoot
Bias & Without & baCART & -0.050 & -0.048 & -0.052 & -0.046 & -0.052 & -0.052 & -0.053 & -0.051 \\
 &  & bCART & -0.054 & -0.053 & -0.056 & -0.047 & -0.054 & -0.055 & -0.055 & -0.054 \\
 & With & baCART & -0.094 & -0.114 & -0.061 & -0.116 & -0.054 & -0.056 & -0.031 & -0.046 \\
 &  & bCART & -0.115 & -0.170 & -0.165 & 0.110 & -0.135 & -0.365 & -0.228 & -0.129 \\
 &  & CCA+baCART & -0.072 & -0.110 & -0.068 & -0.127 & -0.109 & -0.191 & -0.175 & -0.093 \\
 &  & CCA+bCART & -0.063 & -0.070 & -0.040 & -0.104 & -0.102 & -0.169 & -0.171 & -0.079 \\
 &  & MI+baCART & -0.056 & -0.054 & -0.026 & -0.031 & -0.035 & -0.033 & -0.032 & -0.030 \\
 &  & MI+bCART & -0.062 & -0.065 & -0.063 & -0.061 & -0.056 & -0.059 & -0.056 & -0.054 \\
 &  & MI+LRc & -0.010 & -0.014 & -0.006 & 0.000 & -0.006 & -0.005 & -0.004 & -0.004 \\
 &  & MI+LRm & -0.006 & -0.004 & -0.012 & -0.017 & -0.008 & -0.010 & -0.008 & -0.005 
\\ &&&&&&&&&& \\
Empirical & Without & baCART & 0.133 & 0.131 & 0.130 & 0.160 & 0.131 & 0.133 & 0.132 & 0.133 \\
SE &  & bCART & 0.146 & 0.144 & 0.147 & 0.178 & 0.144 & 0.144 & 0.144 & 0.147 \\
 & With & baCART & 0.132 & 0.127 & 0.140 & 0.164 & 0.132 & 0.140 & 0.136 & 0.135 \\
 &  & bCART & 0.148 & 0.143 & 0.134 & 0.165 & 0.142 & 0.139 & 0.151 & 0.145 \\
 &  & CCA+baCART & 0.167 & 0.237 & 0.226 & 0.268 & 0.175 & 0.189 & 0.180 & 0.173 \\
 &  & CCA+bCART & 0.181 & 0.260 & 0.256 & 0.303 & 0.198 & 0.210 & 0.199 & 0.197 \\
 &  & MI+baCART & 0.127 & 0.128 & 0.124 & 0.151 & 0.123 & 0.125 & 0.124 & 0.126 \\
 &  & MI+bCART & 0.139 & 0.138 & 0.133 & 0.162 & 0.131 & 0.133 & 0.132 & 0.133 \\
 &  & MI+LRc & 0.120 & 0.118 & 0.117 & 0.142 & 0.114 & 0.117 & 0.116 & 0.116 \\
 &  & MI+LRm & 0.110 & 0.110 & 0.109 & 0.134 & 0.107 & 0.107 & 0.108 & 0.109 
\\ &&&&&&&&&& \\
Mean & Without & baCART & 0.128 & 0.128 & 0.128 & 0.156 & 0.128 & 0.128 & 0.128 & 0.128 \\
$\widehat{\text{SE}}$ &  & bCART & 0.148 & 0.148 & 0.149 & 0.180 & 0.148 & 0.148 & 0.149 & 0.148 \\
 & With & baCART & 0.128 & 0.126 & 0.127 & 0.153 & 0.128 & 0.127 & 0.129 & 0.127 \\
 &  & bCART & 0.147 & 0.146 & 0.143 & 0.172 & 0.148 & 0.147 & 0.155 & 0.148 \\
 &  & CCA+baCART & 0.160 & 0.237 & 0.224 & 0.262 & 0.173 & 0.182 & 0.174 & 0.173 \\
 &  & CCA+bCART & 0.185 & 0.265 & 0.258 & 0.299 & 0.200 & 0.212 & 0.201 & 0.200 \\
 &  & MI+baCART & 0.140 & 0.146 & 0.139 & 0.166 & 0.139 & 0.139 & 0.139 & 0.139 \\
 &  & MI+bCART & 0.163 & 0.167 & 0.161 & 0.194 & 0.161 & 0.161 & 0.161 & 0.161 \\
 &  & MI+LRc & 0.136 & 0.139 & 0.135 & 0.161 & 0.135 & 0.135 & 0.135 & 0.135 \\
 &  & MI+LRm & 0.124 & 0.127 & 0.124 & 0.150 & 0.123 & 0.124 & 0.123 & 0.123 
\\ &&&&&&&&&& \\
MSE & Without & baCART & 0.020 & 0.019 & 0.020 & 0.028 & 0.020 & 0.020 & 0.020 & 0.020 \\
 &  & bCART & 0.024 & 0.024 & 0.025 & 0.034 & 0.024 & 0.024 & 0.024 & 0.024 \\
 & With & baCART & 0.026 & 0.029 & 0.023 & 0.041 & 0.020 & 0.023 & 0.019 & 0.020 \\
 &  & bCART & 0.035 & 0.049 & 0.045 & 0.039 & 0.038 & 0.152 & 0.074 & 0.038 \\
 &  & CCA+baCART & 0.033 & 0.068 & 0.056 & 0.088 & 0.042 & 0.072 & 0.063 & 0.039 \\
 &  & CCA+bCART & 0.037 & 0.073 & 0.067 & 0.103 & 0.050 & 0.073 & 0.069 & 0.045 \\
 &  & MI+baCART & 0.019 & 0.019 & 0.016 & 0.024 & 0.016 & 0.017 & 0.016 & 0.017 \\
 &  & MI+bCART & 0.023 & 0.023 & 0.022 & 0.030 & 0.020 & 0.021 & 0.020 & 0.021 \\
 &  & MI+LRc & 0.014 & 0.014 & 0.014 & 0.020 & 0.013 & 0.014 & 0.013 & 0.014 \\
 &  & MI+LRm & 0.012 & 0.012 & 0.012 & 0.018 & 0.012 & 0.012 & 0.012 & 0.012 
\\ &&&&&&&&&& \\
Empirical & Without & baCART & 0.860 & 0.871 & 0.867 & 0.876 & 0.866 & 0.856 & 0.858 & 0.861 \\
90\%CI &  & bCART & 0.878 & 0.886 & 0.879 & 0.896 & 0.885 & 0.884 & 0.882 & 0.883 \\
coverage & With & baCART & 0.798 & 0.761 & 0.826 & 0.787 & 0.856 & 0.830 & 0.870 & 0.858 \\
 &  & bCART & 0.795 & 0.685 & 0.707 & 0.840 & 0.773 & 0.188 & 0.569 & 0.775 \\
 &  & CCA+baCART & 0.851 & 0.856 & 0.880 & 0.874 & 0.833 & 0.702 & 0.719 & 0.846 \\
 &  & CCA+bCART & 0.888 & 0.892 & 0.900 & 0.894 & 0.858 & 0.790 & 0.773 & 0.875 \\
 &  & MI+baCART & 0.901 & 0.912 & 0.928 & 0.926 & 0.926 & 0.921 & 0.921 & 0.921 \\
 &  & MI+bCART & 0.920 & 0.927 & 0.935 & 0.942 & 0.934 & 0.935 & 0.937 & 0.934 \\
 &  & MI+LRc & 0.939 & 0.947 & 0.944 & 0.942 & 0.948 & 0.943 & 0.942 & 0.946 \\
 &  & MI+LRm & 0.933 & 0.945 & 0.941 & 0.932 & 0.941 & 0.940 & 0.935 & 0.938 
\\\bottomrule
\end{longtable}
\end{landscape}

\begin{centering}
\begin{table}\centering
\makebox[\textwidth]{
\add{
\begin{tabular}{lld{1.3}d{1.3}d{1.3}d{1.3}d{1.3}d{1.3}}
\toprule
Missing&&\multicolumn{5}{c}{Metric}\\
\cline{3-8}
data&Method&\multicolumn{1}{c}{Bias}&\multicolumn{1}{c}{Bias dif.}&\multicolumn{1}{c}{Emp. SE}&\multicolumn{1}{c}{Mean $\widehat{\text{SE}}$}&\multicolumn{1}{c}{MSE}&\multicolumn{1}{c}{Coverage} \\ \midrule
\multicolumn{8}{c}{\textit{Inverse probability weighting}}\\[2pt]
Without & baCART & 0.061 & \multicolumn{1}{c}{\text{ref.}} & 0.104 & 0.106 & 0.015 & 0.851\\
 & bCART & 0.028 & \multicolumn{1}{c}{\text{ref.}} & 0.119 & 0.123 & 0.015 & 0.902\\
With & baCART & 0.008 & -0.053 & 0.113 & 0.113 & 0.013 & 0.902\\
 & bCART & -0.019 & -0.048 & 0.118 & 0.121 & 0.014 & 0.909\\
 & CCA+baCART & 0.039 & -0.023 & 0.171 & 0.178 & 0.031 & 0.906\\
 & CCA+bCART & 0.041 & 0.013 & 0.185 & 0.192 & 0.036 & 0.903\\
 & MI+baCART & 0.068 & 0.007 & 0.105 & 0.109 & 0.016 & 0.846\\
 & MI+bCART & 0.027 & -0.001 & 0.119 & 0.128 & 0.015 & 0.919\\
 & MI+LRc & 0.004 &  & 0.132 & 0.141 & 0.017 & 0.927\\
 & MI+LRm & 0.005 &  & 0.130 & 0.138 & 0.017 & 0.920
\\ &&&&&& \\
\multicolumn{8}{c}{\textit{Matching}}\\[2pt]
Without & baCART & -0.003 & \multicolumn{1}{c}{\text{ref.}} & 0.121 & 0.121 & 0.015 & 0.901\\
 & bCART & -0.061 & \multicolumn{1}{c}{\text{ref.}} & 0.133 & 0.138 & 0.021 & 0.879\\
With & baCART & -0.045 & -0.042 & 0.124 & 0.121 & 0.017 & 0.869\\
 & bCART & -0.137 & -0.075 & 0.134 & 0.137 & 0.037 & 0.731\\
 & CCA+baCART & -0.097 & -0.094 & 0.220 & 0.222 & 0.058 & 0.871\\
 & CCA+bCART & -0.092 & -0.031 & 0.239 & 0.245 & 0.066 & 0.880\\
 & MI+baCART & 0.007 & 0.010 & 0.117 & 0.134 & 0.014 & 0.938\\
 & MI+bCART & -0.063 & -0.001 & 0.129 & 0.155 & 0.021 & 0.923\\
 & MI+LRc & 0.007 &  & 0.112 & 0.131 & 0.013 & 0.946\\
 & MI+LRm & 0.007 &  & 0.112 & 0.131 & 0.013 & 0.944
\\\bottomrule
\end{tabular}}}
\caption*{\add{Supplementary Table 2: Performance metrics of inverse probability weighting and matching estimators in 5000 simulated datasets of additional simulation experiment. Abbreviations: Bias dif., estimated bias after minus estimated bias before introduction missing data; Emp. SE, empirical standard error; Mean $\widehat{\text{SE}}$, mean estimated standard error; MSE, mean squared error; CART, classification and regression trees; baCART, bootstrap aggregated CART; bCART, boosted CART; CCA, complete case analysis; MI, multiple imputation; LRc, logistic regression with correctly specified model; LRm, logistic regression with main effects only.}}
\end{table}
\end{centering}


\begin{thebibliography}{55}
\newcommand{\enquote}[1]{``#1''}
\providecommand{\natexlab}[1]{#1}
\providecommand{\url}[1]{\texttt{#1}}
\providecommand{\urlprefix}{URL }

\bibitem[{Albert and Anderson(1984)}]{Albert1984}
Albert, A. and J.~Anderson (1984): \enquote{On the existence of maximum
  likelihood estimates in logistic regression models,} \emph{Biometrika}, 71,
  1--10.

\bibitem[{Ali et~al.(2015)Ali, Groenwold, Belitser, Pestman, Hoes, Roes, {de
  Boer}, and Klungel}]{Ali2015}
Ali, M., R.~Groenwold, S.~Belitser, W.~Pestman, A.~Hoes, K.~Roes, A.~{de Boer},
  and O.~Klungel (2015): \enquote{Reporting of covariate selection and balance
  assessment in propensity score analysis is suboptimal: a systematic review,}
  \emph{Journal of Clinical Epidemiology}, 68, 122--131.

\bibitem[{Austin(2011{\natexlab{a}})}]{Austin2011intro}
Austin, P. (2011{\natexlab{a}}): \enquote{An introduction to propensity score
  methods for reducing the effects of confounding in observational studies,}
  \emph{Multivariate Behavioral Research}, 46, 399--424.

\bibitem[{Austin(2011{\natexlab{b}})}]{Austin2011}
Austin, P. (2011{\natexlab{b}}): \enquote{Optimal caliper widths for
  propensity-score matching when estimating differences in means and
  differences in proportions in observational studies,} \emph{Pharmaceutical
  Statistics}, 10, 150--161.

\bibitem[{Austin and Stuart(2015)}]{Austin2015}
Austin, P. and E.~Stuart (2015): \enquote{Moving towards best practice when
  using inverse probability of treatment weighting ({IPTW}) using the
  propensity score to estimate causal treatment effects in observational
  studies,} \emph{Statistics in Medicine}, 34, 3661--3679.

\bibitem[{Belitser et~al.(2011)Belitser, Martens, Pestman, Groenwold, Boer, and
  Klungel}]{Belitser2011}
Belitser, S., E.~Martens, W.~Pestman, R.~Groenwold, A.~Boer, and O.~Klungel
  (2011): \enquote{Measuring balance and model selection in propensity score
  methods,} \emph{Pharmacoepidemiology and Drug Safety}, 20, 1115--1129.

\bibitem[{Breiman(1996)}]{Breiman1996}
Breiman, L. (1996): \enquote{Bagging predictors,} \emph{Machine Learning}, 24,
  123--140.

\bibitem[{Breiman(2001)}]{Breiman2001}
Breiman, L. (2001): \enquote{Random forests,} \emph{Machine Learning}, 45,
  5--32.

\bibitem[{Burgette and Reiter(2010)}]{Burgette2010}
Burgette, L. and J.~Reiter (2010): \enquote{Multiple imputation for missing
  data via sequential regression trees,} \emph{American journal of
  epidemiology}, 172, 1070--1076.

\bibitem[{Cham and West(2016)}]{Cham2016}
Cham, H. and S.~West (2016): \enquote{Propensity score analysis with missing
  data,} \emph{Psychological Methods}, 21, 427--445.

\bibitem[{Cole and Frangakis(2009)}]{Cole2009}
Cole, S. and C.~Frangakis (2009): \enquote{The consistency statement in causal
  inference: a definition or an assumption?} \emph{Epidemiology}, 20, 3--5.

\bibitem[{{D'Agostino Jr.} and Rubin(2000)}]{D'Agostino2000}
{D'Agostino Jr.}, R. and D.~Rubin (2000): \enquote{Estimating and using
  propensity scores with partially missing data,} \emph{Journal of the American
  Statistical Association}, 95, 749--759.

\bibitem[{Doove et~al.(2014)Doove, {van Buuren}, and Dusseldorp}]{Doove2014}
Doove, L., S.~{van Buuren}, and E.~Dusseldorp (2014): \enquote{Recursive
  partitioning for missing data imputation in the presence of interaction
  effects,} \emph{Computational Statistics \& Data Analysis}, 72, 92--104.

\bibitem[{Drake(1993)}]{Drake1993}
Drake, C. (1993): \enquote{Effects of misspecification of the propensity score
  on estimators of treatment effect,} \emph{Biometrics}, 49, 1231--1236.

\bibitem[{Elith et~al.(2008)Elith, Leathwick, and Hastie}]{Elith2008}
Elith, J., J.~Leathwick, and T.~Hastie (2008): \enquote{A working guide to
  boosted regression trees,} \emph{Journal of Animal Ecology}, 77, 802--813.

\bibitem[{Franklin et~al.(2014)Franklin, Rassen, Ackermann, Bartels, and
  Schneeweiss}]{Franklin2014}
Franklin, J., J.~Rassen, D.~Ackermann, D.~Bartels, and S.~Schneeweiss (2014):
  \enquote{Metrics for covariate balance in cohort studies of causal effects,}
  \emph{Statistics in Medicine}, 33, 1685--1699.

\bibitem[{Groenwold et~al.(2009)Groenwold, Nelson, Nichol, Hoes, and
  Hak}]{Groenwold2009}
Groenwold, R., D.~Nelson, K.~Nichol, A.~Hoes, and E.~Hak (2009):
  \enquote{Sensitivity analyses to estimate the potential impact of unmeasured
  confounding in causal research,} \emph{International Journal of
  Epidemiology}, 39, 107--117.

\bibitem[{Groenwold et~al.(2012)Groenwold, White, Donders, Carpenter, Altman,
  and Moons}]{Groenwold2012}
Groenwold, R.~H., I.~R. White, A.~R.~T. Donders, J.~R. Carpenter, D.~G. Altman,
  and K.~G. Moons (2012): \enquote{Missing covariate data in clinical research:
  when and when not to use the missing-indicator method for analysis,}
  \emph{Canadian Medical Association Journal}, 184, 1265--1269.

\bibitem[{Hastie et~al.(2009)Hastie, Tibshirani, and Friedman}]{Hastie2009}
Hastie, T., R.~Tibshirani, and J.~Friedman (2009): \emph{The Elements of
  Statistical Learning: Data Mining, Inference, and Prediction}, New York:
  Springer, second edition.

\bibitem[{Hern\'{a}n and Robins(2017)}]{Hernan2017}
Hern\'{a}n, M. and J.~Robins (2017): \enquote{Fine point 4.3: Collapsibility of
  the odds ratio,} in M.~Hern\'{a}n and J.~Robins, eds., \emph{Causal
  Inference}, Boca Raton: Chapman \& Hall/CRC,
  \urlprefix\url{https://www.hsph.harvard.edu/miguel-hernan/causal-inference-book/},
  forthcoming.

\bibitem[{Holland(1986)}]{Holland1986}
Holland, P. (1986): \enquote{Statistics in causal inference,} \emph{Journal of
  the American Statistical Association}, 81, 945--960.

\bibitem[{Holland(1988)}]{Holland1988}
Holland, P. (1988): \enquote{Causal inference, path analysis, and recursive
  structural equations models,} \emph{Sociological Methodology}, 18, 449--484.

\bibitem[{Lee et~al.(2010)Lee, Lessler, and Stuart}]{Lee2010}
Lee, B., J.~Lessler, and E.~Stuart (2010): \enquote{Improving propensity score
  weighting using machine learning,} \emph{Statistics in Medicine}, 29,
  337--346.

\bibitem[{Lesko et~al.(2017)Lesko, Buchanan, Westreich, Edwards, Hudgens, and
  Cole}]{Lesko2017}
Lesko, C., A.~Buchanan, D.~Westreich, J.~Edwards, M.~Hudgens, and S.~Cole
  (2017): \enquote{Generalizing study results: a potential outcomes
  perspective,} \emph{Epidemiology}, 28, 553--561.

\bibitem[{Leyrat et~al.(2017)Leyrat, Seaman, White, Douglas, Smeeth, Kim,
  Resche-Rigon, Carpenter, and Williamson}]{Leyrat2017}
Leyrat, C., S.~R. Seaman, I.~R. White, I.~Douglas, L.~Smeeth, J.~Kim,
  M.~Resche-Rigon, J.~R. Carpenter, and E.~J. Williamson (2017):
  \enquote{Propensity score analysis with partially observed covariates: How
  should multiple imputation be used?} \emph{Statistical methods in medical
  research}, 0962280217713032.

\bibitem[{Lumley(2014)}]{survey2014}
Lumley, T. (2014): \emph{survey: Analysis of complex survey samples (R package,
  version 3.31)}, Comprehensive R Archive Network, Vienna, Austria,
  \urlprefix\url{http://cran.r-project.org/web/packages/survey/index.html}.

\bibitem[{McCaffrey et~al.(2004)McCaffrey, Ridgeway, and
  Morral}]{McCaffrey2004}
McCaffrey, D., G.~Ridgeway, and A.~Morral (2004): \enquote{Propensity score
  estimation with boosted regression for evaluating adolescent substance abuse
  treatment,} \emph{Psychological Methods}, 9, 403--425.

\bibitem[{Meng(1994)}]{Meng1994}
Meng, X.-L. (1994): \enquote{Multiple-imputation inferences with uncongenial
  sources of input,} \emph{Statistical Science}, 538--558.

\bibitem[{Moisen(2008)}]{Moisen2008}
Moisen, G. (2008): \enquote{Classification and regression trees,} in
  S.~Jorgensen and B.~Fath, eds., \emph{Encyclopedia of Ecology}, volume~1,
  Oxford: Elsevier.

\bibitem[{Moons et~al.(2006)Moons, Donders, Stijnen, and
  Harrell~Jr}]{Moons2006}
Moons, K.~G., R.~A. Donders, T.~Stijnen, and F.~E. Harrell~Jr (2006):
  \enquote{Using the outcome for imputation of missing predictor values was
  preferred,} \emph{Journal of clinical epidemiology}, 59, 1092--1101.

\bibitem[{Neyman et~al.(1935)Neyman, Iwaszkiewicz, and {St.
  Kolodziejczyk}}]{Neyman1935}
Neyman, J., K.~Iwaszkiewicz, and {St. Kolodziejczyk} (1935):
  \enquote{Statistical problems in agricultural experimentation,}
  \emph{Supplement to the Journal of the Royal Statistical Society}, 2,
  107--180.

\bibitem[{Pearl(2009)}]{Pearl2009}
Pearl, J. (2009): \emph{Causality: Models, Reasoning and Inference}, New York:
  Cambridge University Press.

\bibitem[{{Penning de Vries} and Groenwold(2016)}]{PenningdeVries2016}
{Penning de Vries}, B. and R.~Groenwold (2016): \enquote{Comments on propensity
  score matching following multiple imputation,} \emph{Statistical Methods in
  Medical Research}, 25, 3066--3068.

\bibitem[{Peters and Hothorn(2017)}]{ipred2017}
Peters, A. and T.~Hothorn (2017): \emph{ipred: Improved Predictors (R package,
  version 0.9-6)}, Comprehensive R Archive Network, Vienna, Austria,
  \urlprefix\url{http://cran.r-project.org/web/packages/ipred/index.html}.

\bibitem[{{R Core Team}(2016)}]{R2016}
{R Core Team} (2016): \emph{R: A language and environment for statistical
  computing}, R Foundation for Statistical Computing, Vienna, Austria,
  \urlprefix\url{https://www.R-project.org/}.

\bibitem[{Rai et~al.(2017)Rai, Lee, Dalman, Newschaffer, Lewis, and
  Magnusson}]{Rai2017}
Rai, D., B.~Lee, C.~Dalman, C.~Newschaffer, G.~Lewis, and C.~Magnusson (2017):
  \enquote{Antidepressants during pregnancy and autism in offspring: population
  based cohort study,} \emph{BMJ}, 385, j2811.

\bibitem[{Ridgeway(1999)}]{Ridgeway1999}
Ridgeway, G. (1999): \enquote{The state of boosting,} \emph{Computing Science
  and Statistics}, 31, 172--181.

\bibitem[{Ridgeway et~al.(2017)Ridgeway, McCaffrey, Morral, Griffin, and
  Burgette}]{twang2017}
Ridgeway, G., D.~McCaffrey, A.~Morral, B.~Griffin, and L.~Burgette (2017):
  \emph{twang: Toolkit for Weighting and Analysis of Nonequivalent Groups (R
  package, version 1.5)}, Comprehensive R Archive Network, Vienna, Austria,
  \urlprefix\url{http://cran.r project.org/web/packages/twang/index.html}.

\bibitem[{Rosenbaum and Rubin(1983)}]{Rosenbaum1983}
Rosenbaum, P. and D.~Rubin (1983): \enquote{The central role of the propensity
  score in observational studies for causal effects,} \emph{Biometrika}, 70,
  41--55.

\bibitem[{Rubin(1974)}]{Rubin1974}
Rubin, D. (1974): \enquote{Estimating causal effects of treatments in
  randomized and nonrandomized studies,} \emph{Journal of Educational
  Psychology}, 66, 688--701.

\bibitem[{Rubin(1976)}]{Rubin1976}
Rubin, D. (1976): \enquote{Inference and missing data,} \emph{Biometrika}, 63,
  581--592.

\bibitem[{Rubin(1987)}]{Rubin1987}
Rubin, D. (1987): \emph{Multiple imputation for nonresponse in surveys}, New
  York: Wiley.

\bibitem[{Rubin et~al.(2008)}]{Rubin2008}
Rubin, D.~B. et~al. (2008): \enquote{For objective causal inference, design
  trumps analysis,} \emph{The Annals of Applied Statistics}, 2, 808--840.

\bibitem[{Schafer(1997)}]{Schafer1997}
Schafer, J. (1997): \emph{Analysis of incomplete multivariate data}, Boca
  Raton: CRC Press.

\bibitem[{Setoguchi et~al.(2008)Setoguchi, Schneeweiss, MA, Glynn, and
  Cook}]{Setoguchi2008}
Setoguchi, S., S.~Schneeweiss, M.~B. MA, R.~Glynn, and E.~Cook (2008):
  \enquote{Evaluating uses of data mining techniques in propensity score
  estimation: a simulation study,} \emph{Pharmacoepidemiology and Drug Safety},
  17, 546--555.

\bibitem[{Shah(2014)}]{rfimpute}
Shah, A. (2014): \emph{CALIBERrfimpute: Imputation in MICE using Random Forest
  (R package, version 0.1-2)}, Comprehensive R Archive Network, Vienna,
  Austria,
  \urlprefix\url{http://cran.r-project.org/web/packages/CALIBERrfimpute/index.html}.

\bibitem[{Shah et~al.(2014)Shah, Bartlett, Carpenter, Nicholas, and
  Hemingway}]{Shah2014}
Shah, A., J.~Bartlett, J.~Carpenter, O.~Nicholas, and H.~Hemingway (2014):
  \enquote{Comparison of random forest and parametric imputation models for
  imputing missing data using mice: a caliber study,} \emph{American Journal of
  Epidemiology}, 179, 764--774.

\bibitem[{St\"{u}rmer et~al.(2006)St\"{u}rmer, Joshi, Glynn, Avorn, Rothman,
  and Schneeweiss}]{Sturmer2006}
St\"{u}rmer, T., M.~Joshi, R.~Glynn, J.~Avorn, K.~Rothman, and S.~Schneeweiss
  (2006): \enquote{A review of the application of propensity score methods
  yielded increasing use, advantages in specific settings, but not
  substantially different estimates compared with conventional multivariable
  methods. journal of clinical epidemiology,} \emph{Journal of Clinical
  Epidemiology}, 59, 437--e1.

\bibitem[{Tchetgen and VanderWeele(2012)}]{Tchetgen2012}
Tchetgen, E.~T. and T.~VanderWeele (2012): \enquote{On causal inference in the
  presence of interference,} \emph{Statistical Methods in Medical Research},
  21, 55--75.

\bibitem[{Therneau and Atkinson(2017)}]{Therneau2017}
Therneau, T. and E.~Atkinson (2017): \enquote{An introduction to recursive
  partitioning using the {RPART} routines,} \emph{Rochester: Mayo Foundation}.

\bibitem[{{Van Buuren}(2012)}]{Buuren2012}
{Van Buuren}, S. (2012): \emph{Flexible imputation of missing data}, Boca
  Raton: CRC Press.

\bibitem[{{Van Buuren} and {Groothuis-Oudshoorn}(2011)}]{mice2011}
{Van Buuren}, S. and K.~{Groothuis-Oudshoorn} (2011): \enquote{mice:
  Multivariate imputation by chained equations in {R},} \emph{Journal of
  Statistical Software}, 45, 1--67.

\bibitem[{Westreich(2012)}]{Westreich2012}
Westreich, D. (2012): \enquote{Berkson's bias, selection bias, and missing
  data,} \emph{Epidemiology}, 23, 159--164.

\bibitem[{Westreich et~al.(2010)Westreich, Lessler, and {Jonsson
  Funk}}]{Westreich2010}
Westreich, D., J.~Lessler, and M.~{Jonsson Funk} (2010): \enquote{Propensity
  score estimation: neural networks, support vector machines, decision trees
  (cart), and meta-classifiers as alternatives to logistic regression,}
  \emph{Journal of clinical epidemiology}, 63, 826--833.

\bibitem[{Wyss et~al.(2014)Wyss, Ellis, Brookhart, Girman, {Jonsson Funk},
  LoCasale, and St\"{u}rmer}]{Wyss2014}
Wyss, R., A.~Ellis, M.~Brookhart, C.~Girman, M.~{Jonsson Funk}, R.~LoCasale,
  and T.~St\"{u}rmer (2014): \enquote{The role of prediction modeling in
  propensity score estimation: an evaluation of logistic regression, bcart, and
  the covariate-balancing propensity score,} \emph{American Journal of
  Epidemiology}, 180, 645--655.

\end{thebibliography}
\end{document}